\documentclass[review]{fcs}
\usepackage{multirow}
\usepackage{booktabs}
\usepackage{makecell}
\usepackage{threeparttable}
\usepackage{algorithmic}
\usepackage{algorithm}
\usepackage{forest}
\usepackage{changepage}
\usepackage{graphicx}
\usepackage{enumitem}
\usetikzlibrary{arrows.meta,shapes,positioning,shadows,trees}
\usepackage{xcolor}
\definecolor{lightblue}{HTML}{DCE9FF}
\definecolor{darkblue}{HTML}{6C8EBF}
\definecolor{lightgreen}{HTML}{D5E8D4}
\definecolor{darkgreen}{HTML}{82B366}
\definecolor{lightorange}{HTML}{FFE6CC}
\definecolor{darkorange}{HTML}{D79B00}
\definecolor{revised}{RGB}{255,0,0}

\title{A survey on memory-efficient transformer-based model training in AI for science}
\author[1]{Kaiyuan TIAN}
\author*[1]{Linbo QIAO}
\author[1]{Baihui LIU}
\author[1]{Gongqingjian JIANG}
\author[1]{Shanshan LI}
\author*[1]{Dongsheng LI}
\address[1]{College of Computer Science and Technology, National University of Defense Technology, Changsha 410073, China}
\fcssetup{
  received       = {month dd, yyyy},
  accepted       = {month dd, yyyy},
  corr-email     = {qiao.linbo@nudt.edu.cn; dsli@nudt.edu.cn},
}
\begin{abstract}
Scientific research faces high costs and inefficiencies with traditional methods, but the rise of deep learning and large language models (LLMs) offers innovative solutions. This survey reviews transformer-based LLM applications across scientific fields such as biology, medicine, chemistry, and meteorology, underscoring their role in advancing research. However, the continuous expansion of model size has led to significant memory demands, hindering further development and application of LLMs for science. This survey systematically reviews and categorizes memory-efficient pre-training techniques for large-scale transformers, including algorithm-level, system-level, and hardware-software co-optimization. Using AlphaFold 2 as an example, we demonstrate how tailored memory optimization methods can reduce storage needs while preserving prediction accuracy. By bridging model efficiency and scientific application needs, we hope to provide insights for scalable and cost-effective LLM training in AI for science.
\end{abstract}

\begin{document}
\section{Introduction}
The rapid advancement of artificial intelligence, particularly large language models (LLMs), has positioned deep learning (DL) as a vital tool for addressing various scientific challenges. DL offers a new approach to scientific research due to its strengths in automatic feature extraction and complex pattern recognition. According to the universal approximation theorem, deep neural networks (DNNs) can effectively fit any continuous function. Additionally, the high-performance parallel implementation of the backpropagation algorithm on GPUs enables DL to solve scientific and engineering problems efficiently\cite{tang_why_2021a}. DL models demonstrate great potential to assist researchers in accelerating scientific discoveries, enhancing prediction accuracy, and providing innovative solutions to complex challenges.

\begin{figure}[!t]
\centering
\includegraphics[width=\linewidth]{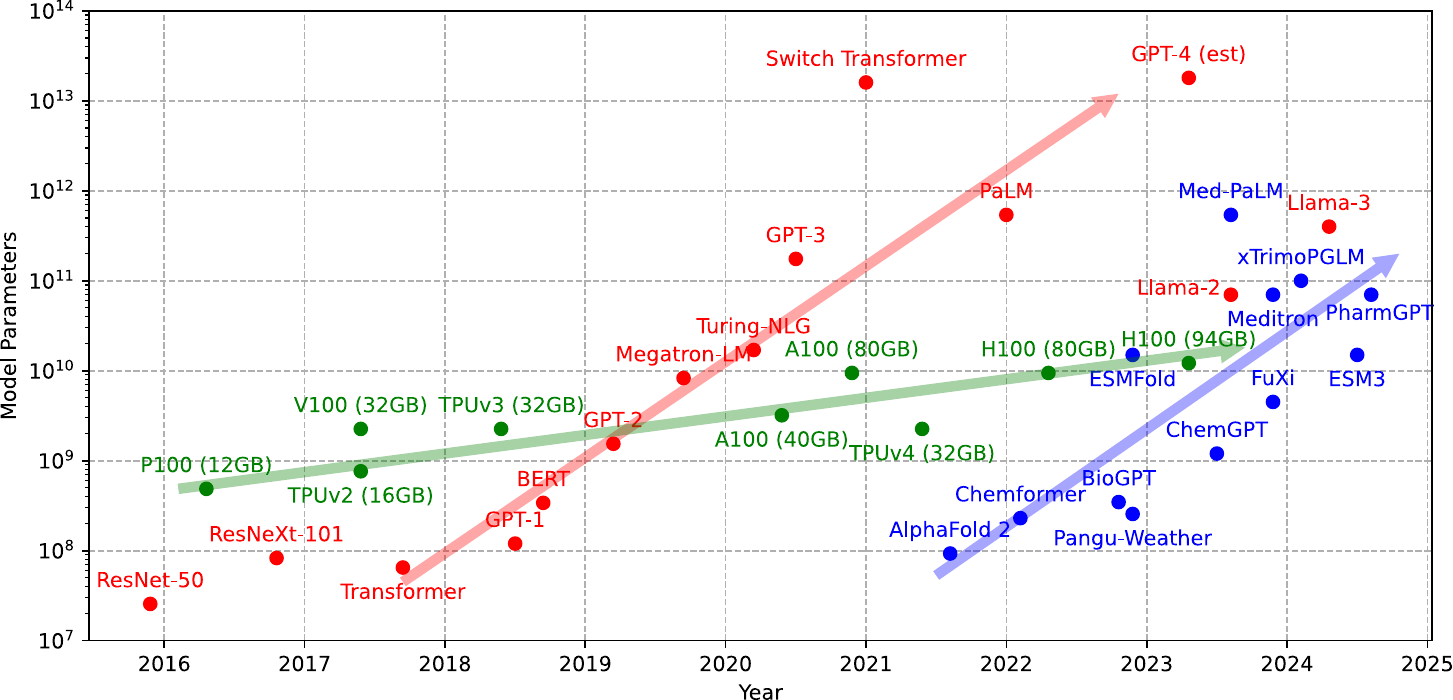}
\caption{AI memory wall. General LLMs are highlighted in red, while LLMs in scientific fields are highlighted in blue. The growth of model parameters has surpassed the increase in memory capacity of accelerators.}
\label{fig:ai_memory_wall}
\end{figure}

In Biological Sciences, DL approaches have proven effective in areas such as protein structure and function prediction. One notable example is AlphaFold 2 \cite{jumper_highly_2021}, which predicts the three-dimensional structures of proteins. In the medical field, LLMs have been applied to various tasks, including medical Q\&A, disease prediction, and clinical decision support. For instance, the Med-PaLM series of models \cite{singhal_large_2023, singhal_expertlevel_2023, tu_generalist_2024} can answer questions posed by medical professionals and assist doctors in diagnosing diseases and determining treatment options. In chemistry, DL is employed for tasks such as molecular generation \cite{frey_neural_2023, fang_domainagnostic_2023}, molecular property prediction \cite{ock_catalyst_2023}, and chemical reaction prediction \cite{irwin_chemformer_2022}. In meteorology, by processing vast amounts of meteorological data, DL models can reveal the complex laws of weather change, improve the accuracy of weather prediction, simulate and predict a wide range of climate patterns \cite{bi_panguweather_2022, bi_accurate_2023, chen_fuxi_2023}, and predict the occurrence and impact of extreme weather events \cite{nguyen_climax_2023, bodnar_aurora_2024}. With the improvement of computing power and DL algorithms, the application of DL in scientific fields will be more extensive, further expanding the boundaries of scientific exploration.

Meanwhile, studies have shown that with parameters increase, the model shows better performance \cite{kaplan_scaling_2020a, hoffmann_empirical_2022}. Consequently, the scale of these models has continued to grow, leading to memory usage issues in DL models. In recent years, the widespread development and application of transformer models \cite{vaswani_attention_2017} have led to a significant increase in model size, averaging a 240-fold growth every two years. The expansion of parameters in models like GPT-3 \cite{brown_language_2020} and Llama 2 \cite{touvron_llama_2023} has outpaced the growth of GPU memory. The training of large-scale models inevitably encounters the ``memory wall'' \cite{gholami_ai_2024} (see Fig. \ref{fig:ai_memory_wall}). Addressing this challenge is critical for enabling the training of large-scale models in resource-constrained environments.

There are a few related surveys summarized memory-efficient LLM training methods. Zhuang et al. \cite{zhuang_survey_2023} have provided an overview of the efficient training technologies used for transformer-based models. They discussed some memory-efficient training methods and summarized the key factors that affect training efficiency. To gain a deeper understanding of the latest advancements in efficient LLMs, Wan et al. \cite{wan_efficient_2024} conducted a thorough literature review and proposed a taxonomy to organize the relevant research.

In contrast to these studies, this survey investigates transformer-based models applied across different scientific fields and their corresponding memory-efficient pre-training techniques. Moreover, we classify them into existing base models, and systematically introduce and categorize trending general memory-efficient pre-training methods, thereby bridging the gap between memory-efficient training methodologies and the need for scaling scientific models. Furthermore, we use a case study to demonstrate studies aimed at optimizing memory overhead in transformer-variants-based models. To the best of our knowledge, there is a notable absence of a comprehensive review that summarizes the application of memory-efficient training techniques within the domain of AI for Science, this survey is the first to systematically connect memory-efficient training techniques to LLMs in AI for science.

Our investigation indicates that despite the widespread application of transformers in scientific fields, there is a notable lack of memory efficiency in their training processes. A judicious selection of memory optimization techniques is essential, as it can significantly improve the efficiency of DL models and broaden their applicability in resource-constrained settings. It is of considerable theoretical and practical importance for facilitating the broader adoption of DL in scientific research.

The remainder of this paper is organized as follows. In Section \ref{sec:Large Language Models and AI for Science}, we begin by providing a brief introduction to deep learning and the associated knowledge of LLMs to enhance understanding for researchers and practitioners from various fields. We then highlight the accomplishments of LLMs across different scientific domains and summarize the memory optimization techniques employed during model pre-training. Section \ref{sec:Memory Efficient Training Techniques of Transformers} offers a comprehensive overview of various strategies aimed at reducing memory overhead. Section \ref{sec:Tailored Memory-Efficient Training Techniques: A Case Study} focuses on AlphaFold 2 as a representative example, detailing some customized memory optimization methods. Section \ref{sec:Future Trends and Challenges} addresses the trends and challenges of applying memory optimization techniques in large-scale models within scientific contexts. Finally, Section \ref{sec:Conclusion} reviews and summarizes the key points discussed throughout the paper.

\section{Large Language Models and AI for Science}\label{sec:Large Language Models and AI for Science}
\subsection{A Brief Introduction to Deep Learning and LLMs}
Deep learning is a machine learning approach that automatically extracts features from complex data through multiple layers of nonlinear transformations. In contrast to traditional machine learning techniques, deep learning offers several advantages, including no need for manual feature engineering, the ability to adapt to complex high-dimensional data, and superior generalization capabilities. In recent years, the emergence of transformer models has further enriched this field.


The proposal of the Transformer architecture \cite{vaswani_attention_2017} marked a revolutionary breakthrough in the field of language modeling. Transformers use a multi-head self-attention mechanism, enabling them to efficiently capture global dependencies in long sequences and enhance model training speed. Models like BERT \cite{devlin_bert_2019a} and GPT \cite{radford_improving_2018a} have emerged one after another, establishing pre-training and fine-tuning as a standard paradigm in natural language processing (NLP). Subsequently, GPT-2 \cite{radford_language_2019} and GPT-3 \cite{brown_language_2020} expanded their parameter sizes to 1.5B and 175B, demonstrating the potential for large-scale pre-training, particularly in zero-shot and few-shot learning. The Scaling Laws \cite{kaplan_scaling_2020a, hoffmann_empirical_2022} reveal the relationship between the growth of model parameters, data size, computational complexity, and model performance, providing theoretical guidance for designing and training large-scale models. To further improve the performance of the model on tasks, researchers have also proposed methods such as instruction fine-tuning \cite{chung_scaling_2024} and InstructGPT \cite{ouyang_training_2022a}. At the end of 2022, ChatGPT showcased the tremendous potential of LLM to the world and sparked a research boom in LLM.

In recent years, companies and research institutions have introduced a range of LLMs, advancing their scale and capabilities. Notable examples include but not limited to Google's PaLM series \cite{chowdhery_palm_2023, anil_palm_2023}, Meta's Llama series \cite{touvron_llama_2023a, touvron_llama_2023, grattafiori_llama_2024}, OpenAI's GPT-4 \cite{openai_gpt4_2024}, DeepSeek-AI's DeepSeek series \cite{deepseek-ai_deepseek_2024, deepseek-ai_deepseekv2_2024, deepseek-ai_deepseekv3_2024}, Tsinghua's GLM-4 \cite{glm_chatglm_2024}, and Alibaba's Qwen series \cite{bai_qwen_2023, yang_qwen2_2024}. So far, LLMs have exhibited exceptional proficiency in language generation, multimodal integration, and specialized tasks, providing substantial support for scientific research, industrial applications, and everyday life. 

\subsection{Application of LLMs in Scientific Fields}
With the success of LLMs in NLP, an increasing amount of research is exploring its application to scientific fields. 
This section will focus on the use of transformer-based LLMs in common tasks across different scientific disciplines as summarized in Table \ref{tab:llm4sci}. Across domains, medical models (e.g., Med-PaLM \cite{singhal_large_2023}, Meditron \cite{chen_meditron70b_2023}) emphasize instruction tuning and expert QA performance; chemistry-oriented models (e.g., Chemformer \cite{irwin_chemformer_2022}, MolGen \cite{fang_domainagnostic_2023}) focus on molecule generation and reaction prediction; climate and Earth science models deal with long-sequence and spatiotemporal data modeling, often requiring architecture adaptation. These models vary in size, training data, and optimization techniques, reflecting distinct memory challenges in each scientific domain task. By reviewing these applications, we aim to illustrate the broad applicability and capabilities of LLMs, as well as the necessity of memory-efficient training techniques. Table \ref{tab:llm4sci} outlines the memory optimization techniques employed during model training. 
It is noteworthy that as activation memory usage varies with experiment settings, the straightforward estimation of memory cost in Table \ref{tab:llm4sci} does not include activations, which constitute a predominant portion of memory overhead during the training process of certain models.

\begin{table*}[!t]
    \caption{Applications of LLMs in scientific fields and corresponding training optimization strategies\label{tab:llm4sci}}
    \centering
    \resizebox{\linewidth}{!}{
    \renewcommand\arraystretch{1.5}
    \begin{threeparttable}
    \begin{tabular}{cccccccc}
        \Xhline{1px}
        \makebox[0.05\textwidth][c]{\textbf{Field}} & \makebox[0.1\textwidth][c]{\textbf{Work}} & \makebox[0.1\textwidth][c]{\textbf{Backbone}} & \makebox[0.1\textwidth][c]{\textbf{Main Building Block}} & \makebox[0.1\textwidth][c]{\textbf{\#Parameters}} & \makebox[0.1\textwidth][c]{\textbf{Memory Cost (est.)}} & \makebox[0.1\textwidth][c]{\textbf{Optimizations}} & \makebox[0.1\textwidth][c]{\textbf{Tasks}}\\
        \hline
        \multirow{15}{*}{\textbf{Biology}} & AlphaFold 2 \cite{jumper_highly_2021} & - & Evoformer \cite{jumper_highly_2021} & 93M & 1.45 GB & \makecell[c]{DP\\mixed-precision\\GC} & protein structure prediction \\ \cline{2-8} 
        {} & RoseTTAFold \cite{baek_accurate_2021} & - & SE(3)-Transformer \cite{fuchs_se3transformers_2020} & 130M & 2.03 GB & \makecell[c]{DP\\GA} & protein structure prediction \\ \cline{2-8} 
        {} & AlphaFold 3 \cite{abramson_accurate_2024a} & - & Pairformer \cite{abramson_accurate_2024a} & Unreported & - & Unreported & \makecell[c]{biomolecular complex\\ structure prediction} \\ \cline{2-8} 
        {} & OpenFold \cite{ahdritz_openfold_2024} & - & Evoformer & 93M & 1.45 GB & \makecell[c]{DP (ZeRO-2)\\mixed-precision\\offloading\\GC, GA} & protein structure prediction \\ \cline{2-8} 
        {} & FastFold \cite{cheng_fastfold_2024} & - & Evoformer & 93M & 1.45 GB & \makecell[c]{DP\\DAP \cite{cheng_fastfold_2024}} & protein structure prediction \\ \cline{2-8} 
        {} & ScaleFold \cite{zhu_scalefold_2024a} & - & Evoformer & 97M & 1.52 GB & \makecell[c]{DP\\DAP} & protein structure prediction \\ \cline{2-8} 
        {} & ESMFold \cite{lin_evolutionaryscale_2023} & ESM-2 15B & Transformer \cite{vaswani_attention_2017} & 15B & 240 GB & DP (FSDP \cite{zhaoyanli_pytorch_2023}) & protein structure prediction \\ \cline{2-8} 
        {} & xTrimoPGLM \cite{chen_xtrimopglm_2024} & xTrimoPGLM-100B & Transformer & 100B & 1.56 TB & \makecell[c]{DP (ZeRO-1)\\PP (1F1B)\\TP (Megatron-LM)\\mixed-precision\\GC} & \makecell[c]{protein understanding\\protein generation} \\ \cline{2-8} 
        {} & ESM3 \cite{hayes_simulating_2024} & - & Transformer & 1.4B / 7B / 98B & 1.53 TB & \makecell[c]{DP (FSDP)\\mixed-precision} & \makecell[c]{protein reasoning\\protein generation} \\ 
        \hline
        \multirow{13}{*}{\textbf{Medicine}} & BioGPT \cite{luo_biogpt_2022} & GPT-2 Medium \cite{radford_language_2019} & Transformer & 347M & 5.42 GB & \makecell[c]{DP\\GA} & \makecell[c]{biomedical text understanding\\biomedical text generation} \\ \cline{2-8}
        {} & Med-PaLM \cite{singhal_large_2023} & \makecell[c]{PaLM 540B \cite{chowdhery_palm_2023}\\Flan-PaLM 540B \cite{chung_scaling_2024}} & Transformer & 540B & 8.44 TB & \makecell[c]{DP (ZeRO-3)\\TP\\GC} & \makecell[c]{medical question answering\\medical reasoning} \\ \cline{2-8}
        {} & Med-PaLM 2 \cite{singhal_expertlevel_2023} & PaLM 2 340B \cite{anil_palm_2023} & Transformer & 340B & 5.31 TB & Unreported & \makecell[c]{medical question answering\\medical reasoning} \\ \cline{2-8}
        {} & Med-PaLM M \cite{tu_generalist_2024} & PaLM-E \cite{driess_palme_2023} & Transformer & 12B / 84B / 562B & 8.78 TB & \makecell[c]{DP (ZeRO-3)\\TP\\GC} & biomedical generalist \\ \cline{2-8}
        {} & BiomedGPT \cite{zhang_generalist_2024} & OFA \cite{wang_ofa_2022} & Transformer & 33M / 93M / 182M & 2.84 GB & \makecell[c]{DP (PyTorch DDP)\\mixed-precision} & biomedical generalist \\ \cline{2-8}
        {} & Meditron \cite{chen_meditron70b_2023} & Llama-2 \cite{touvron_llama_2023} & Transformer & 7B / 70B & 1.09 TB & \makecell[c]{DP, PP\\TP (Megatron-LM)} & medical reasoning \\ \cline{2-8}
        {} & HuatuoGPT \cite{zhang_huatuogpt_2023a} & \makecell[c]{Baichuan-7B\\Ziya-LLaMA-13B \cite{zhang_fengshenbang_2023}} & Transformer & 7B / 13B & 208 GB & DP (ZeRO) & medical consultation \\ \cline{2-8}
        {} & HuatuoGPT-\uppercase\expandafter{\romannumeral2} \cite{chen_huatuogptii_2024a} & \makecell[c]{Baichuan2 \cite{yang_baichuan_2023}\\Yi-34B \cite{ai_yi_2024}} & Transformer & 7B / 13B / 34B & 544 GB & DP (ZeRO) & medical consultation \\ 
        \hline
        \multirow{2.4}{*}{\textbf{Biomedicine}} & PharmBERT \cite{valizadehaslani_pharmbert_2023} & BERT-Base \cite{devlin_bert_2019a} & Transformer & 110M & 1.72 GB & Unreported & drug labeling \\ \cline{2-8}
        {} & PharmGPT \cite{chen_pharmagpt_2024} & Llama-2 & Transformer & 3B / 13B / 70B & 1.09 TB & \makecell[c]{DP+PP+TP} & \makecell[c]{text understanding\\text generation} \\
        \hline
        \multirow{7}{*}{\textbf{Chemistry}} & ChemBERT \cite{guo_automated_2022} & BERT-Base & Transformer & 110M & 1.72 GB & Unreported & \makecell[c]{product extraction\\reaction role labeling} \\ \cline{2-8}
        {} & CatBERTa \cite{ock_catalyst_2023} & RoBERTa \cite{liu_roberta_2019} & Transformer & 355M & 5.55 GB & Unreported & catalyst property prediction \\ \cline{2-8}
        {} & Chemformer \cite{irwin_chemformer_2022} & BART \cite{lewis_bart_2020} & Transformer & 45M / 230M & 3.59 GB & \makecell[c]{DP (ZeRO-2)\\mixed-precision} & \makecell[c]{reaction prediction\\molecular optimization\\molecular property prediction} \\ \cline{2-8}
        {} & MolGen \cite{fang_domainagnostic_2023} & BART & Transformer & 355M & 5.55 GB & \makecell[c]{DP (ZeRO-2)\\mixed-precision\\GA} & molecule generation \\ \cline{2-8}
        {} & ChemGPT \cite{frey_neural_2023} & GPT-Neo \cite{black_gptneo_2021} & Transformer & 1.2B & 19.20 GB & DP (PyTorch DDP) & molecule generation \\ 
        \hline
        \multirow{6.3}{*}{\textbf{Meteorology}} & Pangu-Weather \cite{bi_accurate_2023} & - & Swin transformer \cite{liu_swin_2021} & 256M & 4 GB & DP & weather forecast \\ \cline{2-8}
        {} & FuXi \cite{chen_fuxi_2023} & - & Swin transformer \cite{liu_swin_2022} & 4.5B & 72 GB & \makecell[c]{DP (FSDP)\\mixed-precision\\GC} & weather forecast \\ \cline{2-8}
        {} & ClimaX \cite{nguyen_climax_2023} & - & ViT \cite{dosovitskiy_image_2021} & Unreported & - & \makecell[c]{DP\\mixed-precision} & \makecell[c]{weather forecast\\climate projection} \\ \cline{2-8}
        {} & Aurora \cite{bodnar_aurora_2024} & - & Swin transformer & 1.3B & 20.80 GB & \makecell[c]{DP\\mixed-precision\\GC} & atmospheric prediction \\ 
        \hline
        \multirow{4.1}{*}{\textbf{Geoscience}} & K2 \cite{deng_k2_2024} & Llama-7B \cite{touvron_llama_2023} & Transformer & 7B & 112 GB & \makecell[c]{DP (ZeRO-3)\\mixed-precision\\GA} & foundational model \\ \cline{2-8}
        {} & GeoGalactica \cite{lin_geogalactica_2024} & Galactica-30B \cite{taylor_galactica_2022} & Transformer& 30B & 480 GB & DP+PP+TP+SP & foundational model \\ \cline{2-8}
        {} & JiuZhou \cite{chen_jiuzhou_2025} & \makecell[c]{Mistral-7B \cite{jiang_mistral_2023}\\Qwen1.5-14B \cite{bai_qwen_2023}} & Transformer & 7B / 14B & 224 GB & \makecell[c]{DP (ZeRO-2)\\mixed-precision\\GA} & foundational model \\ \cline{2-8}
        \Xhline{1px}
    \end{tabular}
    \begin{tablenotes}
        \footnotesize
        \item DP: data parallelism, ~~TP: tensor parallelism, ~~PP: pipeline parallelism, ~~SP: sequence parallelism, ~~GA: gradient accumulation, ~~GC: gradient checkpointing
    \end{tablenotes}
    \end{threeparttable}
    }
\end{table*}

\subsubsection{Biology}
Protein structure prediction represents a significant challenges in biochemistry, as high-precision predictions are essential for drug discovery and protein design. AlphaFold 2 \cite{jumper_highly_2021} is a DL-based tool for predicting protein structures. It can predict the three-dimensional structure of proteins based on their amino acid sequences.

The architecture of AlphaFold 2 is built upon variants of Transformer. Its main structure features an input module that receives the amino acid sequence of the target protein and searches for homologs within a sequence database to perform multiple sequence alignment (MSA). The Evoformer module then processes the MSA representation and pair representation, facilitating the exchange and updating of information through Transformer layers. Finally, the structural module converts the abstract representation of proteins into 3D atomic coordinates \cite{yang_alphafold2_2023}. AlphaFold 2 has significantly influenced the advancement of biotechnology. However, its training process and datasets are not fully open source. 
Furthermore, despite the AlphaFold 2 model having a relatively small parameter size (about 93M), the memory consumed by intermediate activations is substantial due to its architecture, and peak memory usage increases cubically with the length of the input sequence. 

Although AlphaFold2 and RoseTTAFold \cite{baek_accurate_2021} achieved breakthroughs in atomic resolution structure prediction, they rely on MSA and similar protein structure templates to achieve better performance. Based on the research of OpenFold \cite{ahdritz_openfold_2024}, Lin et al. \cite{lin_evolutionaryscale_2023} further explored the ability of language model and proposed ESMFold. It utilizes the internal representation of language model to replace the expensive network module for processing MSA with a transformer module for processing sequences. This simplification greatly improves the speed of ESMFold, which is much higher than MSA-based models. 

The protein language model (PLM) has achieved remarkable success in learning biological information from protein sequences. Most existing models are constrained by autoencoding or autoregressive pre-training objectives, making it challenging to address both protein understanding and generation tasks. Chen et al. \cite{chen_xtrimopglm_2024} explored the possibility of jointly optimizing these two training objectives through an innovative framework and trained a 100B PLM (xTrimoPGLM). Inspired by ESMFold, they combined the folding module with xTrimoPGLM and developed xT-Fold, a high-performance 3D structure prediction tool. The performance of xT-Fold is superior to other PLM-based models on benchmarks such as CAMEO and CASP15. 

In May 2024, AlphaFold 3 was released \cite{abramson_accurate_2024a}. The overall architecture of Alphafold 3 is similar to AlphaFold 2, yet it demonstrates enhanced prediction accuracy and efficiency. dditionally, the application scope of Alphafold 3 has been expanded. It can predict the structure of a single protein, the complex structure of protein complexes, proteins, nucleic acids, small molecules, etc.

\subsubsection{Medicine}
LLMs have demonstrated significant potential in the medical and clinical fields. 
The applications of such medical models are numerous, including medical knowledge retrieval, assistance in clinical diagnoses, consultations, and personalized medical services.

Singhal et al. \cite{singhal_large_2023} found that the benchmark used to evaluate the clinical knowledge of the model is limited. To solve it, they proposed MultiMedQA, a benchmark composed of seven medical Q\&A datasets. Further, they fine-tuned the command of PaLM \cite{chowdhery_palm_2023} and got Flan-PaLM \cite{chung_scaling_2024}, which shows excellent performance in multiple-choice questions. Moreover, they suggested instruction prompt tuning, and the obtained Med-PaLM model can further adapt to medical field. 

Tu et al. \cite{tu_generalist_2024} aim to further explore the potential of generalist models (those capable of performing multiple tasks without fine-tuning, such as GPT-3, PaLM, and GPT-4) in medical field. Utilizing the language and multimodal foundation models PaLM 540B and PaLM-E 562B \cite{driess_palme_2023}, the researchers develop Med-PaLM M. They also designed a multimodal biomedical benchmark called MultiMedBench. As a biomedical generalist model, Med-PaLM M demonstrates strong performance across all tasks on MultiMedBench, matching or exceeding the capabilities of task-specific models.

Currently, most LLMs utilized in medical field are either closed source or have relatively small model sizes. Chen et al. \cite{chen_meditron70b_2023} believe that while models designed for generalist tasks typically perform well across tasks, their performance in certain tasks falls short compared to models tailored for those specific purposes. They developed Meditron-70B through continued pre-training of Llama-2 and fine-tuned it for particular tasks, resulting in superior performance compared to all open-source generalist and medical LLMs across all medical benchmarks. 

HuatuoGPT \cite{zhang_huatuogpt_2023a} aims to assist doctors with diagnosis and treatment in Chinese scenarios, it combines the dialogue capabilities of ChatGPT with the expertise of real-world physicians. By employing instruction fine-tuning and reinforcement learning techniques, it exploits distilled instruction data generated by ChatGPT alongside actual response data from doctors. The upgraded version, HuatuoGPT-\uppercase\expandafter{\romannumeral2} \cite{chen_huatuogptii_2024a}, successfully passed the Chinese National Pharmacist Licensure Examination in October 2023 and nearly all medical qualification exams, highlighting its robust capabilities within the Chinese medical landscape. 

\subsubsection{Biomedicine}
The emergence and rapid development of LLM heralded a new era in the fields of biopharmaceuticals and chemical sciences, providing innovative methods for drug discovery, chemical synthesis and optimization, and elucidation of complex biological pathways.

Drug labels possess distinct characteristics that differentiate them from texts in other domains, making it ineffective to apply generic models directly to this specific field. ValizadehAslani et al. \cite{valizadehaslani_pharmbert_2023} leveraged drug label text corpora and conducted pre-training on BERT base checkpoints using masked language modeling (MLM) as the pre-training task. This effort resulted in PharmBERT, a domain-specific BERT model for drug labeling. Their work showcased exceptional performance by evaluating PharmBERT on three downstream tasks.

To tackle the issue of limited knowledge among general-purpose language models in the domains of biopharmaceuticals and chemistry, Chen et al. introduced PharmaGPT \cite{chen_pharmagpt_2024}, a large-scale multilingual language model tailored for these fields. PharmaGPT is pre-trained on high-quality, large-scale, domain-specific datasets and optimized for enhanced performance in biopharmaceuticals and chemistry through fine-tuning and reinforcement learning. Experiments indicate that PharmaGPT outperforms mainstream general models on professional benchmark tests, such as NAPLEX, showcasing its exceptional capability in understanding and generating specialized knowledge.

\subsubsection{Chemistry}
The application of LLM in the field of chemistry covers multiple aspects, including molecular design, molecular property prediction, and chemical reaction prediction. Additionally, LLMs can be used to analyze and summarize chemical literature, providing researchers with efficient knowledge retrieval and decision-making support. 

Extracting structured data from chemical literature is a challenging task due to the complexity and heterogeneity of chemical language. Guo et al. \cite{guo_automated_2022} broke this task into two cascaded subtasks: product extraction and reaction role labeling. They developed two independent models, ChemBERT and ChemRxnBERT, based on BERT. By pre-training and fine-tuning these models on domain-specific and task-oriented corpora, they enhance the understanding of semantic relationships and contextual features in chemical texts.

In the field of molecular property prediction, Ock et al. introduced CatBERTa \cite{ock_catalyst_2023}, which can bypass the requirement of precise atomic coordinates in previous GNN-based methods and predict catalyst properties sorely through molecular text representation. Compared to graph-based methods, using text representation to describe adsorbate-catalyst systems has better interpretability. The prediction accuracy of CatBERTa is comparable to GNN-based methods, highlighting its potential in property prediction tasks.

Irwin et al. pre-trained BART \cite{lewis_bart_2020} on a dataset containing a large number of simplified molecular line entry system (SMILES) strings to capture the structural relationships and potential properties between chemical molecules, resulting in the Chemformer model \cite{irwin_chemformer_2022}. Chemformer is capable of both sequence-to-sequence tasks (e.g. reaction prediction, molecular optimization), and discriminative tasks (e.g. molecular property prediction, biological activity analysis). It achieves or exceeds the performance of state-of-the-art (SOTA) models on multiple benchmarks while significantly reducing convergence time. 

The objective of molecular design and generation tasks is to create new molecules with specific functions or properties. 
Traditional methodologies depend heavily on the expertise of specialists and costly computational simulations. 
Fang et al. introduced a molecular generative pre-trained language model called MolGen \cite{fang_domainagnostic_2023}, which has enhanced its comprehension of chemical structures and semantics through a two-stage pre-training strategy. This strategy encompasses molecular syntax and semantic learning, utilizing self-sacrificing embedded strings along with domain-agnostic prefix tuning. 
Frey et al. \cite{frey_neural_2023} conducted an in-depth study on the neural scaling behavior of deep chemical models in relation to dataset and model size, aiming to explore how model performance is enhanced with increased resource investment. They designed and trained a generative pre-trained Transformer model named ChemGPT with over 1B parameters for small molecule generation. By varying both the model and dataset sizes by several orders of magnitude, they demonstrated an improvement in pre-training loss performance as the scale increased.

\subsubsection{Meteorology}
Over the past decade, the advancement of high-performance computing devices has led to significant progress in numerical weather prediction (NWP), enhancing the accuracy of forecasts for daily weather, extreme weather events, and climate change. However, as the growth of computing power slows and the complexity of physical models increases, the limitations of traditional methods have become more pronounced. Recently, the rapid progress in DL has emerged as a promising avenue for addressing these challenges.

The Pangu-Weather model \cite{bi_panguweather_2022, bi_accurate_2023}, proposed by Bi et al., employs the 3D Earth-Specific Transformer (3DEST) architecture to process 3D climate data, which is a variant of Swin Transformer \cite{liu_swin_2021, liu_swin_2022}. It is the first AI-driven approach to surpass the accuracy of NWP methods. Notably, Pangu-Weather achieves an inference speed of 1.4 seconds on a single GPU, making it 10000× faster than the previous most accurate NWP method (operational IFS \cite{wedi_modelling_2015}). While the accuracy of 3D models can surpass 2D models, this enhancement comes with a considerable increase in memory requirements, leading to limited network depth. 

The inherent uncertainty in weather forecasting is inevitable, and the degree of this uncertainty, along with cumulative errors, increase as the forecast period lengthens. In response to this challenge, Chen et al. introduced FuXi \cite{chen_fuxi_2023}, a cascaded model architecture based on U-Transformer, which effectively reduce cumulative errors and improves both forecast accuracy and duration. The model pre-training was conducted on a cluster of 8 Nvidia A100 GPUs.

Inspired by the concept of foundational models \cite{bommasani_opportunities_2022}, Nguyen et al. designed and trained a basic model named ClimaX \cite{nguyen_climax_2023}, aimed at effectively adapting to general tasks related to the Earth's atmosphere. ClimaX can be trained on heterogeneous datasets to solve diverse climate and weather tasks. Remarkably, even with pre-training conducted at lower resolutions and with limited computational resources, ClimaX surpasses previous data-driven models in weather and climate prediction benchmarks. The pre-training used 80 NVIDIA V100 32GB GPUs. 
Subsequently, Bodnar et al. introduced Aurora \cite{bodnar_aurora_2024}, a atmospheric foundation model designed to excel in various prediction tasks, particularly in regions with limited data or extreme weather conditions. After training on diverse weather and climate datasets, Aurora can effectively adapt to new atmospheric prediction tasks by learning a universal representation of atmospheric dynamics. The model was trained on 32 GPUs with 1 batch size per GPU. 

\subsubsection{Geoscience}
As LLMs demonstrate success in multiple domains, expanding their capabilities to Earth science becomes a natural progression. The geoscience field, with its extensive and complex datasets, demands efficient and domain-aware LLM solutions.

K2 \cite{deng_k2_2024} pioneers the application of LLMs in geoscience by extending LLaMA-7B through domain-specific pre-training and expert instruction tuning using the GeoSignal dataset. 
Evaluations on the GeoBench benchmark reveal K2's superior performance over general LLMs in both factual recall and reasoning tasks within the geoscience domain.

Building on this trend, GeoGalactica \cite{lin_geogalactica_2024} further scales geoscience LLMs by specializing in a 30B-parameter model through continued pre-training of Galactica \cite{taylor_galactica_2022} on 65B tokens, followed by instruction fine-tuning. The training relies on Megatron-LM framework and utilizes a computing cluster consisting of 512 nodes, with each node equipped with 4 hygon DCU accelerator. GeoGalactica achieves SOTA performance on various geoscience benchmarks and expert-validated tasks, underscoring the importance of domain-adaptive scaling and memory-efficient training for advancing AI-driven Earth system research.

Complementing these efforts, JiuZhou \cite{chen_jiuzhou_2025} introduces a geoscience foundation model developed through a two-stage pre-adaptation pre-training method on the large-scale JiuZhou-Corpus. 
Despite utilizing a limited two-GPU setup, JiuZhou effectively scales training with long-context inputs and large batch sizes, achieving competitive performance on the GeoBench benchmark and outperforming GPT-3.5 in various geoscientific tasks. This work further illustrates the growing importance of memory-efficient training strategies for advancing LLMs in Earth science.

\subsubsection{Memory Challenges in Scaling Transformers for Scientific Applications}
Despite the progress achieved in applying transformer models across various domains, the continuous increase in memory overhead during model training remains a pressing concern that requires attention. We identify the underlying causes of this issue from the following key factors:
\begin{itemize}[wide]
    \item \textbf{The number of model parameters continues to increase.} In recent years, there has been a notable increase in the scale of models across various fields, as shown in Fig. \ref{fig:ai_memory_wall}. This trend is driven by the demand for processing complex and large-scale data, advancements in computing resources, and the pursuit of enhanced performance and accuracy in downstream tasks. Consequently, this growth has also resulted in higher memory requirements.
    \item \textbf{The necessity to process long sequence inputs.} Long sequence data is prevalent across various scientific domains, such as DNA and protein sequences in biology, molecular structures in chemistry, and time-series weather data in meteorology. As the length of input data increases, the memory required to store activations also grows significantly. This challenge is exacerbated by the quadratic computational complexity inherent in transformer models. The situation may become worse when employing a more complex architecture.
    \item \textbf{The demand for handling diverse data structures.} In scientific fields, the use of high-resolution images and multimodal data also significantly contributes to increased activation memory consumption during large model training.
\end{itemize}

\tikzset{
    basic/.style  = {draw, align=center, minimum height=0.66cm, rounded corners=2pt},
    root/.style   = {basic, thin, text width=3.6cm, line width=0.6pt},
    midnode/.style = {basic, thin, fill=lightblue, text width=3.6cm, line width=0.6pt},
    endnode/.style = {basic, thin, align=left, text width=12cm, draw=darkblue, line width=0.8pt},
    algorithm/.style = {basic, thin, fill=lightblue, text width=3.6cm, line width=0.6pt},
    system/.style = {basic, thin, fill=lightgreen, text width=3.6cm, line width=0.6pt},
    hardware/.style = {basic, thin, fill=lightorange, text width=3.6cm, line width=0.6pt},
    algorithm-end/.style = {basic, thin, align=left, text width=10cm, draw=darkblue, line width=0.8pt},
    system-end/.style = {basic, thin, align=left, text width=10cm, draw=darkgreen, line width=0.8pt},
    hardware-end/.style = {basic, thin, align=left, text width=10cm, draw=darkorange, line width=0.8pt},
    edge from parent/.style={draw=black, edge from parent fork right}
}
\begin{figure*}[!t]
    \begin{adjustwidth}{}{}
    \centering
    \resizebox{0.9\linewidth}{!}{
    \begin{forest} for tree={
        anchor=center,
        grow=east,
        growth parent anchor=east,
        parent anchor=east,
        child anchor=west,
        l sep=5mm,  
        reversed,
        edge path={\noexpand\path[\forestoption{edge}, -, line width=0.6pt, >={latex}] 
             (!u.parent anchor) -- +(8pt,0pt) |- (.child anchor)
             \forestoption{edge label};},
    }
    [Memory-Efficient Training Technique, root, rotate=90, parent anchor=south
        [Algorithm (\S\ref{sec:Algorithm}), algorithm
            [Compression (\S\ref{sec:Compression}), algorithm
                [Mixed Precision Training \cite{micikevicius_mixed_2018, kalamkar_study_2019a, micikevicius_fp8_2022, peng_fp8lm_2023}, algorithm, fill=white, draw=darkblue, line width=0.8pt]
                [Quantization-Aware Training, algorithm
                    [{LLM-QAT \cite{liu_llmqat_2024}, BitNet \cite{wang_bitnet_2023}, BitNet b1.58 \cite{ma_era_2024}, EfficientQAT \cite{chen_efficientqat_2024}}, algorithm-end]
                ]
            ]
            [Memory Efficient Optimizers (\S\ref{sec:Memory Efficient Optimizers}), algorithm
                [First-Order, algorithm
                    [{Adafactor \cite{shazeer_adafactor_2018}, SM3 \cite{anil_memory_2019}, CAME \cite{luo_came_2023b}, Lion \cite{chen_symbolic_2023a}, Adam-mini \cite{zhang_adammini_2024}}, algorithm-end] ]
                [Zeroth-Order, algorithm
                    [{MeZO \cite{malladi_finetuning_2023}, DeepZero \cite{chen_deepzero_2023}, ZO-AdaMU \cite{jiang_zoadamu_2024}}, algorithm-end] ]
            ]
            [Gradient Checkpointing (\S\ref{sec:Gradient Checkpointing}), algorithm
                [{Full Checkpointing \cite{chen_training_2016}, Selective Checkpointing \cite{korthikanti_reducing_2023}}, algorithm-end] 
            ]
            [Gradient Accumulation (\S\ref{sec:Gradient Accumulation}), algorithm]
        ]
        [System (\S\ref{sec:System}), system
            [Distributed Training\\(\S\ref{sec:Distributed Training}), system
                [Data Parallelism, system
                    [{Pytorch DDP \cite{li_pytorch_2020}, ZeRO \cite{rajbhandari_zero_2020b}, FSDP \cite{zhaoyanli_pytorch_2023}}, system-end, text width=12cm] ]
                [Tensor Parallelism, system
                    [{Megatron-LM \cite{shoeybi_megatronlm_2020}, 2D-TP \cite{xu_efficient_2023a}, 2.5D-TP \cite{wang_tesseract_2022}, 3D-TP \cite{bian_maximizing_2021}}, system-end, text width=12cm]]
                [Pipeline Parallelism, system
                    [{GPipe \cite{huang_gpipe_2019}, 1F1B \cite{harlap_pipedream_2018, narayanan_efficient_2021a, fan_dapple_2021a}, Interleaved 1F1B \cite{narayanan_efficient_2021a}, Zero Bubble \cite{qi_zero_2024, qi_pipeline_2024}}, system-end, text width=12cm]]
                [Sequence Parallelism, system
                    [{Ring Self-Attention \cite{li_sequence_2023}, Megatron SP \cite{korthikanti_reducing_2023}, Ulysses \cite{jacobs_system_2024}, Ring Attention \cite{liu_ringattention_2023}}, system-end, text width=12cm] ]
            ]
            [Offloading (\S\ref{sec:Offloading}), system
                [{SwapAdvisor \cite{huang_swapadvisor_2020}, ZeRO-Offload \cite{ren_zerooffload_2021}, ZeRO-Infinity \cite{rajbhandari_zeroinfinity_2021}}, system-end]
            ]
        ]
        [Hardware-Software Co-Optimization (\S\ref{sec:Hardware}), hardware
            [{FlashAttention \cite{dao_flashattention_2022}, FlashAttention-2 \cite{dao_flashattention2_2023}, FlashAttention-3 \cite{shah_flashattention3_2024}, MPress \cite{zhou_mpress_2023}}, hardware-end, text width=14.35cm]
        ]
    ]
    \end{forest}
    }
    \end{adjustwidth}
    \caption{Overview of memory-efficient training techniques for transformers.}
    \label{fig:overview}
\end{figure*}

\section{Memory-Efficient Training Techniques of Transformers}\label{sec:Memory Efficient Training Techniques of Transformers}
LLMs are evolving rapidly and find extensive applications across various domains. However, the training of these models imposes significant demands on computational resources and memory. In this context, exploring memory-efficient training strategies is critical to enabling broader accessibility and supporting larger model sizes. 
While LLMs have gained considerable popularity in scientific fields and have led to noteworthy results, our analysis suggests that many existing works employ only a limited subset of available memory optimization techniques. As shown in Table \ref{tab:llm4sci}, most works adopt one or more common memory-saving techniques (e.g., mixed precision training, DP), but more advanced strategies remain underutilized. This may limit the ability to train larger or more efficient models. In the rest of this section, we will delve into various memory-efficient training techniques specific to transformer-based models, as illustrated in Fig. \ref{fig:overview}.

\subsection{Algorithm}\label{sec:Algorithm}
A variety of algorithmic techniques have been proposed to address memory challenges. These techniques optimize memory usage while maintaining training efficiency and model accuracy. This section categorizes and analyzes algorithmic strategies for memory optimization in transformer models. Table \ref{tab:overview-algorithm-level} provides an overview of various algorithm-level memory-efficient training methods, summarizing their characteristics and typical use cases.

\subsubsection{Compression}\label{sec:Compression}
Compression techniques reduce memory usage by minimizing the precision or redundancy of data representations during training, while aiming to retain comparable performance. These methods are widely used in resource-constrained scenarios.

\begin{enumerate}[label=\roman*., listparindent=-\labelwidth, wide]
    \item \textbf{Mixed Precision Training}

    Mixed Precision Training leverages data types of varying precisions to reduce memory consumption and computational costs, while preserving comparable accuracy. It was introduced by Micikevicius et al. \cite{micikevicius_mixed_2018} that model parameters, activations, and gradients are stored using FP16 while maintaining a copy of the FP32 parameters. To cope with numerical stability issues caused by reduced numerical range and accuracy, FP16 parameters are used for forward and backward calculations, while FP32 parameters are used for model updates. Additionally, to mitigate the risk of gradient underflow during backpropagation, loss scaling is commonly applied, which helps to minimize model performance degradation. The mixed precision training process is illustrated in Fig. \ref{fig:mix-precision}.

    \begin{figure}[!t]  
    \centering
    \includegraphics[width=0.48\textwidth]{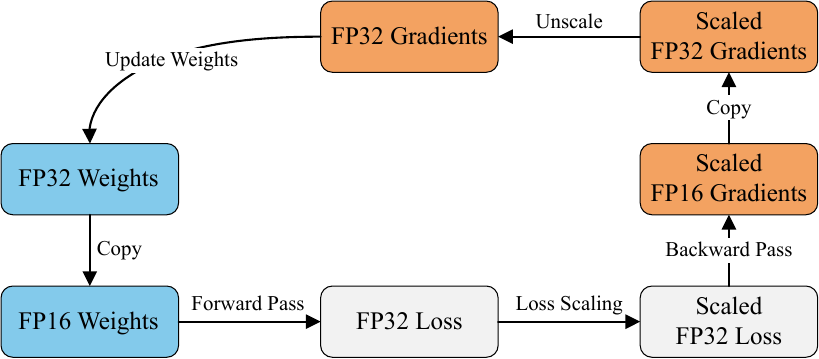}
    \caption{FP16 mixed precision training process.}
    \label{fig:mix-precision}
    \end{figure}
    
    While it needs to save an additional FP32 model parameter, the memory savings achieved by utilizing FP16 activations surpass the memory costs associated with these extra FP32 parameters, leading to an overall memory usage reduction.

    To address the challenges posed by FP16, Google has introduced a new floating-point format known as BF16. Although BF16 offers lower accuracy than FP16, it has a larger numerical range, effectively mitigating numerical overflow during training. This improvement not only enhances computation speed but also accelerates model convergence \cite{kalamkar_study_2019a}. Although some large models are trained with FP16 mixed precision training, the training report of Bloom-176B \cite{scao_bloom_2023a} suggests that BF16 is a superior alternative that helps avoid numerical issues.
    
    Modern GPUs, such as NVIDIA's Volta, Turing, and Ampere architectures, feature specialized Tensor Cores designed to accelerate mixed-precision matrix operations. However, BF16 support is limited to certain hardware platforms, such as Ampere and Hopper architectures. Since the Hopper architecture, FP8 Tensor Cores were added allowing for FP8 mixed precision training, which can further reduce memory consumption and accelerate computation \cite{micikevicius_fp8_2022, peng_fp8lm_2023}.
    
    Mixed precision training requires careful management to prevent numerical issues. As hardware and software continue to evolve, the methods and tools for mixed precision training are also advancing, facilitating the training of larger and more intricate models.
    
    \item \textbf{Quantization-Aware Training}

    Quantization-aware training (QAT) reduces memory consumption and computational demands by representing model parameters and activations in lower-precision formats (e.g., INT8 or FP8) while maintaining accuracy through training-time adaptation. Unlike post-training quantization (PTQ), QAT simulates quantization effects during forward and backward passes, enabling the model to learn robust representations resilient to precision loss.

    Liu et al. introduced LLM-QAT \cite{liu_llmqat_2024}, a QAT framework for LLMs that requires no access to the original pre-training data. They generated synthetic data by sampling next-token outputs from the full-precision pre-trained model and trained quantized model via logit-based knowledge distillation. Applied to LLaMA-7B, 13B, and 30B, LLM-QAT achieves accurate 4-bit weight quantization and consistently outperforms SOTA PTQ methods, especially at precision lower than 8-bit. LLM-QAT can achieve 7.3× speedup in inference with 4-bit weight 8-bit activation quantization with comparable accuracy.

    BitNet \cite{wang_bitnet_2023} replaced standard linear layers with BitLinear modules using 1-bit weights and quantized activations. BitNet was trained via QAT and achieved competitive accuracy with significantly reduced memory and energy consumption compared to FP16 Transformers. The authors demonstrated that BitNet obeys a scaling law akin to full-precision Transformers, enabling efficient scaling to larger models while maintaining training stability and inference performance. Building upon this foundation, Ma et al. proposed BitNet b1.58 \cite{ma_era_2024}, a ternary-weight version (with weights in {-1, 0, +1}) that stroke a more favorable trade-off between accuracy and efficiency. They showed that b1.58-bit models matched or exceeded the performance of FP16 LLaMA models starting from 3B parameters while offering up to 4.1× decoding latency speedup, 7.2× memory savings, and 71.4× energy efficiency. Together, these works highlight the promise of low-bit quantization for scaling and deploying LLMs in a highly resource-efficient manner.

    EfficientQAT \cite{chen_efficientqat_2024} aims to enhance memory and computational efficiency for LLMs while minimizing accuracy loss. It operates in two phases: Block-wise Training of All Parameters (Block-AP) and End-to-End Training of Quantization Parameters (E2E-QP). Experiment results showed that 2-bit EfficientQAT on Llama-2-70B completed training in 41h on a single A100-80GB GPU with peak memory $\approx$ 34 GB, achieving less than 3 points accuracy degradation compared to the FP16 model.

\end{enumerate}

\begin{table*}[t]
\centering
\caption{An overview of algorithm-level memory-efficient training methods}
\label{tab:overview-algorithm-level}
\resizebox{\linewidth}{!}{
    \begin{tabular}{ccll}
    \Xhline{1px}
    \textbf{Category} & \textbf{Method} & \multicolumn{1}{c}{\textbf{Feature}} & \multicolumn{1}{c}{\textbf{Applicable Scenarios}} \\ \hline
    \multirow{4}{*}{Compression} & \begin{tabular}[c]{@{}c@{}}Mixed Precision\\ Training\end{tabular} & \begin{tabular}[c]{@{}l@{}}- Roughly halves overall memory consumption\\ - Speedup training\\ - May cause numerical issues\end{tabular} & - Widely applicable \\ \cline{2-4} 
     & \begin{tabular}[c]{@{}c@{}}Quantization-Aware\\ Training\end{tabular} & \begin{tabular}[c]{@{}l@{}}- Further compresses model size during training\\ - Prepares models for low-bit inference\\ - Potential loss in model accuracy\end{tabular} & \begin{tabular}[c]{@{}l@{}}- Edge deployment scenarios\\ - Resource-constrained training\\ - When quantization compatibility is required\end{tabular} \\ \hline
    \multirow{3}{*}{\begin{tabular}[c]{@{}c@{}}Memory Efficient\\ Optimizers\end{tabular}} & First-Order & \begin{tabular}[c]{@{}l@{}}- Fast convergence\\ - High memory demand\end{tabular} & - Pre-training and finetuning \\ \cline{2-4} 
     & Zeroth-Order & \begin{tabular}[c]{@{}l@{}}- Low memory footprint\\ - Slow convergence\end{tabular} & \begin{tabular}[c]{@{}l@{}}- Finetune on downstream tasks\\ - Resource constrained settings\end{tabular} \\ \hline
    \multirow{3}{*}{\begin{tabular}[c]{@{}c@{}}Gradient Checkpointing\end{tabular}} & Full Checkpointing & - Trades additional computation for reduced memory & \multirow{3}{*}{\begin{tabular}[c]{@{}l@{}}- Avoid if memory is sufficient, as it decreases\\ ~~throughput\end{tabular}} \\ \cline{2-3}
     & Selective Checkpointing & \begin{tabular}[c]{@{}l@{}}- Perform recompute on components that cost substantial\\ ~~memory yet require less computation\end{tabular} &  \\ \hline
    Gradient Accumulation & - & \begin{tabular}[c]{@{}l@{}}- Increase effective batch size\\ - Delay weight updates across multiple steps\end{tabular} & - Useful in low-memory environments \\
    \Xhline{1px}
    \end{tabular}
}
\end{table*}

\subsubsection{Memory Efficient Optimizers}\label{sec:Memory Efficient Optimizers}
Training LLMs often involve maintaining not only model parameters but also optimizer states such as momentum and variance estimates. These auxiliary states can triple the memory footprint, especially when using adaptive optimizers like Adam \cite{kingma_adam_2017} or AdamW \cite{loshchilov_decoupled_2018}, which are widely adopted due to their fast convergence and training stability. However, this comes at a steep memory cost. Memory-efficient optimizers aim to reduce this overhead by redesigning the way these statistics are stored or approximated. These optimizers can be particularly beneficial in scientific applications where training resources are limited but model complexity is high.

Shazeer et al. introduced Adafactor \cite{shazeer_adafactor_2018}, which significantly lowers memory requirements while delivering training performance comparable to Adam. Assuming the model parameter matrix is denoted as $W \in R^{m \times n}$, by applying low-rank decomposition 
on the second-moment estimates of the gradients, Adafactor can reduce memory consumption from $O(m \cdot n)$ to $O(m + n)$. Meanwhile, Anil et al. proposed SM3 \cite{anil_memory_2019}, which reduces the memory requirements for storing optimizer states through a parameter cover mechanism. It involves dividing parameters into multiple subsets, with each subset maintaining only one variable to approximate the second-order statistics of all parameters within it. SM3 employs a data-driven approach to dynamically adjust the learning rate, similar to Adagrad, while mitigating the memory overhead that arises from maintaining parameter-level statistics, as seen in Adagrad. Experiments demonstrate that the training performance of SM3 is on par with that of Adagrad and Adam, while its memory overhead is slightly lower than that of Adafactor.

Adafactor uses non-negative matrix factorization to approximate the second-order statistics of gradients. While it saves memory, it can introduce errors that lead to instability during training. SM3 encounters similar challenges. To tackle this issue, Luo et al. introduced CAME optimizer \cite{luo_came_2023b}. It utilizes the residual between the exponential moving average (EMA) of the update and the current update to compute a confidence matrix, subsequently adjusting parameter updates. By applying non-negative matrix factorization to the confidence matrix, CAME further decreases memory overhead. Through confidence-guided strategies and non-negative matrix factorization, CAME significantly reduces memory usage while ensuring convergence. In LLM training tasks, CAME has shown superior performance compared to Adafactor, while being comparable to it in terms of memory efficiency. 

Chen et al. proposed an approach that treats algorithm discovery as a program search, which was successfully applied to identify optimization algorithms for training deep neural networks, thus discovering the Lion optimizer \cite{chen_symbolic_2023a}. Unlike most adaptive optimizers, Lion solely tracks momentum and employs sign operations to calculate updates, leading to reduced memory overhead and a consistent update magnitude. In comparison to Adam, AdamW, and Adafactor, Lion exhibits outstanding performance across multiple tasks and models. Notably, in language modeling tasks, Lion surpasses Adafactor, particularly in scenarios of large-batch training.

Zhang et al. discovered that the Hessian matrix of the Transformer exhibits an approximate block diagonal structure. Based on this finding, they proposed Adam-mini \cite{zhang_adammini_2024}, which partitions the parameters of components such as Q, K, V, and MLP into blocks and assigns a learning rate to each block, reducing the resource requirement by 90\% to 99\%. In the pre-training tasks of TinyLlama-1B, Llama2-7B, and the GPT series, Adam-mini demonstrated a significant reduction in memory overhead compared to AdamW, achieving savings of 45\% to 50\%. Additionally, it offered faster convergence speed while maintaining comparable or improved performance, which surpassed that of Adafactor, SM3, and CAME. 

Traditional gradient-based optimizers update model parameters through gradients calculated during backpropagation, inevitably leading to additional memory overhead and computational complexity. On the other hand, zeroth-order optimizers estimate gradients based on function values, which can effectively reduce memory usage during model training. MeZO \cite{malladi_finetuning_2023} was the first to use a zeroth-order optimizer for fine-tuning memory optimization of LLMs. By introducing a memory-efficient zeroth-order optimizer, the fine-tuning of LLM can be achieved with only forward passes, significantly lowering memory demands. As a comparison, MeZO can train a 3B model on a single A100 80GB GPU, while traditional optimizers can only train a 270M model under the same hardware budget.

DeepZero \cite{chen_deepzero_2023} primarily addresses the scalability issue of zeroth-order optimization in the training of DNNs. It highlights for the first time the benefits of coordinate gradient estimation compared to random vector gradient estimation. To enhance the deployment and application of zeroth-order optimizer training, DeepZero has introduced methods such as feature reuse and forward parallelization. On the other hand, Zo-AdaMU \cite{jiang_zoadamu_2024} handles the problems of oscillation, overfitting, and slow convergence in MeZO by incorporating momentum and uncertainty into simulated perturbations to refine gradient estimation. It also dynamically adjusts uncertainty and momentum parameters using a simulated annealing mechanism, thereby improving the stability and convergence speed of the optimization process.

\subsubsection{Gradient Checkpointing}\label{sec:Gradient Checkpointing}
\begin{figure}[!t]
\centering
\subfloat[w/o GC]{\includegraphics[width=0.48\textwidth]{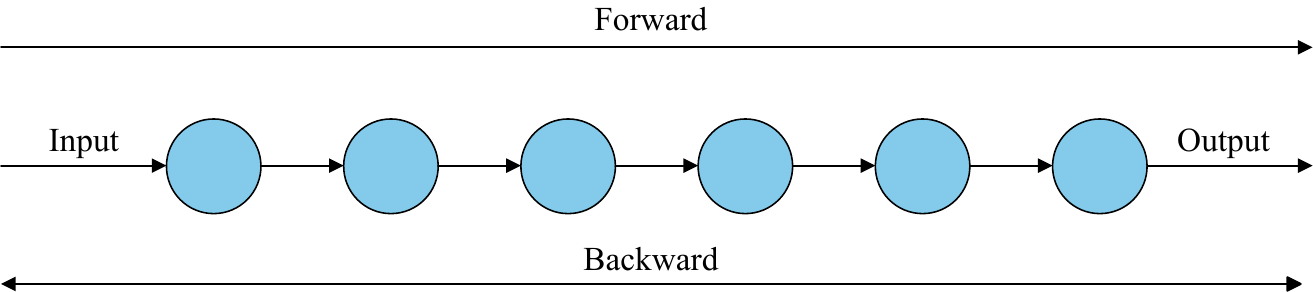}}\label{fig:ckpt_1} \\
\subfloat[w/ GC]{\includegraphics[width=0.48\textwidth]{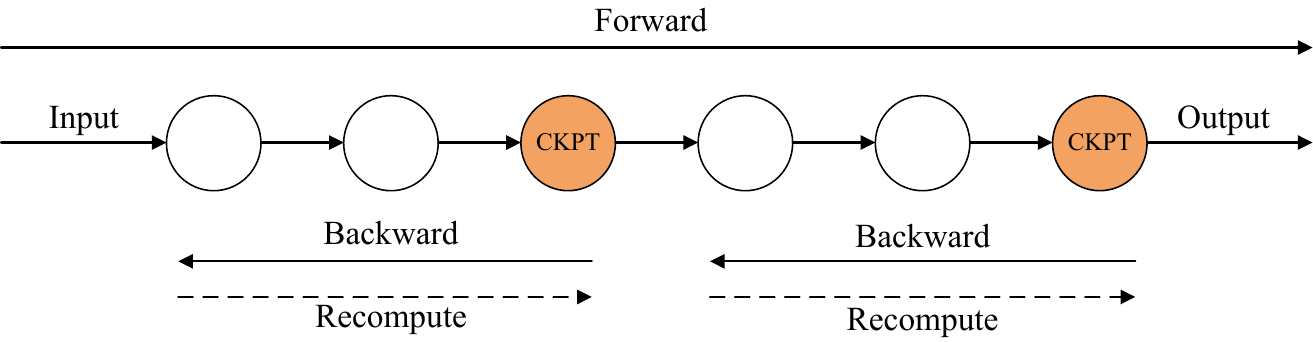}}\label{fig:ckpt_2}
\caption{Comparison of training w/ and w/o gradient checkpointing (GC).}
\label{fig:ckpt}
\end{figure}

During the training process of DNNs, the output of each layer (activations) needs to be stored in memory to calculate gradients during backpropagation. When dealing with large-scale transformers, long input sequences, or large batch sizes, the memory overhead caused by activations can become substantial. Gradient checkpointing reduces memory usage by retaining only the intermediate activations of certain layers during forward propagation while recalculating the discarded activations when needed during backpropagation. The core idea behind it is to trade computation for memory. Fig. \ref{fig:ckpt} shows the difference in whether gradient checkpointing is enabled during the training process. The preserved intermediate activations are referred to as ``checkpoints'' \cite{chen_training_2016}.

Gradient checkpointing offers an effective solution to alleviate memory bottlenecks in the training of deep learning models, particularly in memory-constrained environments. It is important to choose checkpoints carefully to achieve an optimal balance between memory usage and computation time. Taking the transformer architecture as an example, it is generally advisable to set a checkpoint every 1 to 2 transformer layers \cite{narayanan_efficient_2021a}.

The number of parameters and computational complexity of various components in DNNs often vary. Korthikanti et al. \cite{korthikanti_reducing_2023} suggest that for transformer models, it is unnecessary to set checkpoints for the entire transformer layer for recomputation. Instead, selective recomputation can be performed by identifying components of the transformer layer where activations cost substantial memory yet require less computation (e.g., the computation of $Q \times K^T$, subsequent softmax, dropout, and attention score multiplies $V$). By setting checkpoints and recalculating selectively, memory consumption can be optimized without incurring significant computational overhead.

\begin{algorithm}[t]  
\caption{Gradient Accumulation.}\label{alg:Gradient Accumulation}
\begin{algorithmic}[1]
    \REQUIRE{Samples $X$, Labels $Y$, Number of epochs $E$, Iterations per epoch $I$, Gradient accumulation steps $S$}
    \FOR{$i$ from 1 to $E$}
        \FOR{$j$ from 1 to $I$}
            \STATE $output \leftarrow model(X_j)$
            \STATE{$loss \leftarrow \frac{loss\_fn(output_j, Y_j)}{S}$~~// Normalize loss}
            \STATE{$loss.backward()$}
            \IF{$j \bmod S = 0$}
                \STATE{$optimizer.step()$}
                \STATE{$optimizer.zero\_grad()$}
            \ENDIF
        \ENDFOR
    \ENDFOR
\end{algorithmic}
\end{algorithm}

\subsubsection{Gradient Accumulation}\label{sec:Gradient Accumulation}
When training large models, using a larger batch size is often desirable as it can improve training stability, convergence efficiency, and hardware utilization. However, GPU memory limitations may become a bottleneck.
Gradient accumulation is helpful when GPU memory is insufficient to process large batches directly. The fundamental idea behind gradient accumulation is to gradually accumulate gradients from multiple small batches until a desired batch size is achieved, at which point a weight update is performed. For instance, if the total batch size we aim to use is 32, which exceeds the GPU memory capacity, we can divide it into smaller batches with batch sizes of 8, and perform 4 training iterations before conducting a weight update. Algorithm \ref{alg:Gradient Accumulation} presents the pseudocode of gradient accumulation, outlining the implementation of this technique in the training process. 

\subsubsection{Approximate Attention}\label{sec:Approximate Attention}
While most memory-efficient techniques discussed in this survey focus on training-level optimizations, another important class of approaches aims to reduce memory complexity at the architectural level.

Linformer \cite{wang_linformer_2020} reduces attention complexity from $O(n^2)$ to $O(n)$ by projecting the key and value matrices into a lower-dimensional space using learned linear projections. This low-rank approximation significantly reduces both memory and compute cost with on-par model quality.

Performer \cite{choromanski_rethinking_2020} approximates softmax attention through kernel methods, replacing the softmax operation with a feature map transformation. It introduces the FAVOR+ (Fast Attention Via positive Orthogonal Random features) mechanism, enabling linear-time and linear-memory attention. The method maintains high accuracy while improving scalability to longer contexts.

These architectural modifications offer an orthogonal path to memory efficiency compared to the training-time techniques. Although such approximate attention models have not yet been widely applied in AI for Science tasks, they hold strong potential for future adoption. Since the primary focus of this survey is on training-time memory optimization, we only provide a brief overview of these methods here.

\subsection{System}\label{sec:System}
System-level techniques address memory limitations by leveraging inter-device communication and heterogeneous memory hierarchies. This section reviews representative system-level methods that support the training of transformer models through coordinated memory management across computing infrastructure.

\subsubsection{Distributed Training}\label{sec:Distributed Training}
As deep learning models become larger in parameter and data size, it is becoming more challenging for a single computing device to fulfill training needs. Consequently, distributed training has emerged as one of the foundational technologies in deep learning. It involves distributing the training tasks across multiple devices (such as GPUs or nodes) for parallel processing. By employing effective task decomposition and data partitioning strategies, distributed training significantly enhances the efficiency and scalability of model training.

\begin{figure}[t]
    \centering
    \subfloat[Parameter Server]{\includegraphics[width=.43\columnwidth]{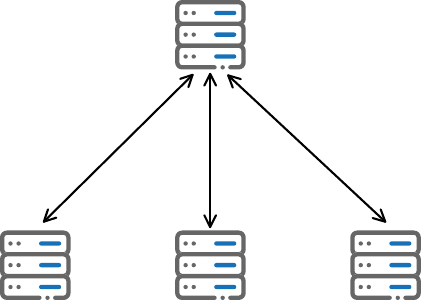}
        \label{fig:PS-architecture}}
    \hspace{5pt}
    \subfloat[Decentralized]{\includegraphics[width=.43\columnwidth]{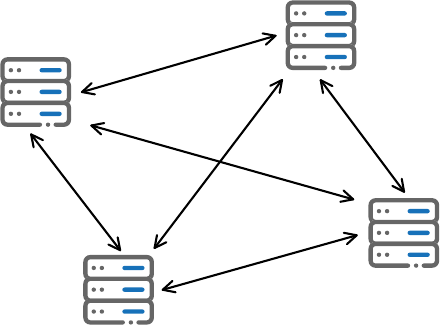}
        \label{fig:decentralized-architecture}}
    \caption{Different architectures for data parallelism.}
    \label{fig:dp-architecture}
\end{figure}

As the scale of models continues to increase, a single GPU's memory can no longer accommodate complete model parameters, optimizer states, and intermediate activations. Traditional training strategies for large-scale models may take weeks or even months. However, distributed training allows the model to be divided across multiple devices, enabling the training of extremely large models with limited single GPU memory and significantly reducing training time. Table \ref{tab:overview-distributed training} provides a brief overview of the features, applicable scenarios, and performance of different distributed training techniques discussed in this section.

\begin{table*}[!t]
\centering
\caption{An overview of distributed training methods}
\label{tab:overview-distributed training}
\resizebox{\linewidth}{!}{
    \begin{tabular}{cclll}
    \Xhline{1px}
     & \textbf{Method} & \multicolumn{1}{c}{\textbf{Feature}} & \multicolumn{1}{c}{\textbf{Applicable Scenarios}} & \multicolumn{1}{c}{\textbf{Performance}} \\ \hline
    \multirow{9}{*}{DP} & DDP & \begin{tabular}[c]{@{}l@{}}- Accelerate training with larger global batch size\\ - moderate communication, can overlap with computation\\ - high redundancy\end{tabular} & \begin{tabular}[c]{@{}l@{}}- Train models that fits in single GPU\\ ~~(e.g., \textless 1B params)\\ - Scale training to more GPUs / nodes\end{tabular} & \begin{tabular}[c]{@{}l@{}}Training throughput scales linearly as the number of GPUs\\  increases\end{tabular} \\ \cline{2-5} 
     & ZeRO-1 & \begin{tabular}[c]{@{}l@{}}- partitioned optimizer states\\ - no extra communication\\ - redundant parameters and gradients\end{tabular} & \multirow{3}{*}{\begin{tabular}[c]{@{}l@{}}- Choose appropriate method according\\ ~~to hardware environment\\ - Train larger models with comparable\\ ~~throughput\\ - Combine with TP/PP to further expand\\ ~~model size\end{tabular}} & 4× memory redunction \\ \cline{2-3} \cline{5-5} 
     & ZeRO-2 & \begin{tabular}[c]{@{}l@{}}- partitioned optimizer states and gradients\\ - no extra communication\\ - redundant parameters\end{tabular} &  & 8× memory reduction \\ \cline{2-3} \cline{5-5} 
     & ZeRO-3 & \begin{tabular}[c]{@{}l@{}}- no redundancy\\ - 1.5× communication\end{tabular} &  & $N_d$× memory reduction \\ \hline
    \multirow{7}{*}{TP} & \begin{tabular}[c]{@{}c@{}}Megatron-LM\\ (1D-TP)\end{tabular} & \begin{tabular}[c]{@{}l@{}}- Split matrix multiplication to reduce computation and\\ ~~parameters stored on each device\\ - Frequent communication\\ - Redundant input activations\end{tabular} & \begin{tabular}[c]{@{}l@{}}- High-speed bandwidth between adjacent GPUs\\ ~~(e.g., NVLink)\\ - Apply within nodes to prevent communication\\ ~~bottleneck\end{tabular} & $N_d$× memory reduction, 77\% throughput (8 GPUs) \\ \cline{2-5} 
     & 2D-TP & \multirow{5}{*}{\begin{tabular}[c]{@{}l@{}}- Lower communication volume\\ - More process groups\\ - Split input data\end{tabular}} & \multirow{5}{*}{\begin{tabular}[c]{@{}l@{}}- GPUs interconnected with different\\ ~~bandwidth\\ - Number of GPU \textgreater{}= 4\end{tabular}} & \begin{tabular}[c]{@{}l@{}}70\% memory consumption when training with 4 GPUs\\  (compared to 1D-TP)\end{tabular} \\ \cline{2-2} \cline{5-5} 
     & 2.5D-TP &  &  & \begin{tabular}[c]{@{}l@{}}56\% memory consumption when training with 8 GPUs\\  (compared to 1D-TP)\end{tabular} \\ \cline{2-2} \cline{5-5} 
     & 3D-TP &  &  & \begin{tabular}[c]{@{}l@{}}35\% memory consumption when training with 8 GPUs\\  (compared to 1D-TP)\end{tabular} \\ \hline
    \multirow{7}{*}{PP} & GPipe & \begin{tabular}[c]{@{}l@{}}- Partition model layers\\ - High peak memory usage\end{tabular} & \multirow{7}{*}{\begin{tabular}[c]{@{}l@{}}- Apply PP across nodes to scale up to larger\\ ~~models\\ - Select methods based on actual needs\end{tabular}} & \begin{tabular}[c]{@{}l@{}}$N_d$× memory reduction, 3.5× throughput\\  (compared to vanilla PP)\end{tabular} \\ \cline{2-3} \cline{5-5} 
     & 1F1B & - Lower peak memory usage &  & \begin{tabular}[c]{@{}l@{}}Reduced activation memory usage with comparable throughput\\ ~~(compared to GPipe)\end{tabular} \\ \cline{2-3} \cline{5-5} 
     & Interleaved 1F1B & \begin{tabular}[c]{@{}l@{}}- Finer-grained scheduling to further reduce bubbles\\ - More communication and activation memory usage\end{tabular} &  & \begin{tabular}[c]{@{}l@{}}Throughput increased by 15\%  with 1.23× memory\\  (compared to 1F1B)\end{tabular} \\ \cline{2-3} \cline{5-5} 
     & Zero Bubble & \begin{tabular}[c]{@{}l@{}}- Split backward pass\\ - Minimum bubbles\end{tabular} &  & \begin{tabular}[c]{@{}l@{}}Throughput increased by 11\%  with 1.08× memory\\  (compared to 1F1B)\end{tabular} \\ \hline
    \multirow{6}{*}{SP} & RSA & - Ring-style calculation of self-attention & - Incompatible with memory-efficient attention & \begin{tabular}[c]{@{}l@{}}71\% memory consumption, 97\% throughput\\  (compared to 1D-TP, sequence length = 2048)\end{tabular} \\ \cline{2-5} 
     & Megatron SP & - Divide layernorm and droupout & - Apply in conjunction with Megatron-TP & \begin{tabular}[c]{@{}l@{}}67\% memory consumption, 6\% speedup\\  (compared to 1D-TP)\end{tabular} \\ \cline{2-5} 
     & Ulysses & - Utilize All-to-All communication to obtain attention heads & \begin{tabular}[c]{@{}l@{}}- Number of attention heads is an integer\\ ~~multiple of SP degree\end{tabular} & \begin{tabular}[c]{@{}l@{}}1.51× throughput vs. Megatron SP,  1.99× throughput vs. RSA\\  (sequence length = 8k, 32 GPUs)\end{tabular} \\ \cline{2-5} 
     & Ring Attention & \begin{tabular}[c]{@{}l@{}}- Ring-style attention computation\\ - Communication can overlap with computation\end{tabular} & - Applicable for both training and inference & \begin{tabular}[c]{@{}l@{}}Enable training large models (7B-65B) on long input (over 4M)\\  with comparable model FLOPS utilization\end{tabular} \\
    \Xhline{1px}
    \end{tabular}
}
\end{table*}

\begin{enumerate}[label=\roman*., listparindent=-\labelwidth, wide]
\item \textbf{Data Parallelism}

Data parallelism (DP) is a commonly used distributed training technique, particularly in environments with multiple GPUs and nodes. DP can significantly accelerate training process and improve overall training throughput. 
It involves replicating the model across computing devices. During training, each device is assigned a subset of the batched training data, known as a mini-batch, which is used as input for the model. Due to the different input data across model replicas, it is necessary to aggregate the gradients from all computing devices before updating model parameters. In terms of implementation, DP architectures can be categorized into parameter server architecture (Fig. \ref{fig:PS-architecture}) and decentralized architecture (Fig. \ref{fig:decentralized-architecture}). In parameter server architecture \cite{li_scaling_2014a}, devices are divided into parameter servers and workers. The parameter server stores the latest parameters of the model and aggregates gradients collected from workers to perform parameter updates. Each worker communicates with the parameter server during training to retrieve the latest parameters or upload gradients calculated using mini-batches.

Parameter servers are likely to become a communication bottleneck, as all workers need to communicate with servers frequently, especially in large-scale training scenarios. In contrast, decentralized architecture avoids this issue. In such a setup, computing device peers perform gradient aggregation through collective communications such as all-reduce, which enhances scalability. Moreover, the implementation of PyTorch DDP \cite{li_pytorch_2020} leverages the overlap between computation and communication to further increase training throughput.

DP divides the input training data along the batch dimension, which reduces the size of intermediate activations on each device during training and greatly improves training efficiency. However, its problem is also obvious. DP requires each device to hold a model replica, which is infeasible for large models that a single device cannot store.

The memory overhead during model training primarily comes from model states and intermediate activations. The model states can be further divided into optimizer states, gradients, and parameters. Assuming the number of parameters in the model is denoted as $P$, and utilizing Adam optimizer with FP16 mixed precision during training, the memory overhead of model parameters is $2P+4P=6P$ bytes. The memory overhead of gradients is $2P$ bytes, and optimizer states consume $4P+4P=8P$ bytes. Thus, the total memory overhead is $16P$ bytes. If training GPT-3 (175B), would require approximately 2.8TB of memory. To address the redundancy of model states, Rajbhandari et al. introduced Zero Redundancy Optimizer (ZeRO) \cite{rajbhandari_zero_2020b}, which shards redundant model states across devices, reducing memory overhead up to $\frac{1}{N_d}$, where $N_d$ represents the number of devices. The core principle of ZeRO is that each device maintains only a fraction of the model states instead of a complete copy. The implementation of ZeRO has been integrated into Microsoft’s open-source distributed training framework, DeepSpeed \cite{rasley_deepspeed_2020a}, making it easily accessible for researchers and developers.

Inspired by ZeRO, PyTorch developed Fully Sharded Data Parallel (FSDP) \cite{zhaoyanli_pytorch_2023}. It demonstrates performance comparable to that of DeepSpeed ZeRO, offering a robust solution for scaling deep learning models. Since PyTorch v2.4
, the PyTorch team has restructured FSDP and introduced FSDP2, adding additional optimizations to enhance performance.

\item \textbf{Tensor Parallelism}

The rapid increase in model parameters has rendered the memory and computing resources of a single device insufficient to support the training of the entire model. Although ZeRO can partition the model across computing devices, it does not reduce computational workloads on each device. For large models with vast parameters and high computational demands, relying solely on DP becomes inadequate for meeting the training needs. Model parallelism (MP) has emerged for further scaling of models. MP primarily includes two techniques: tensor parallelism (TP) and pipeline parallelism (PP), which will be discussed in this section and the following section. TP, also referred to as intra-layer parallelism, finely divides the computation within each model layer into smaller tensor operations that can be executed concurrently across multiple devices, thereby effectively distributing both memory overhead and computational workload.

\begin{figure}[t]  
\centering
\includegraphics[width=0.48\textwidth]{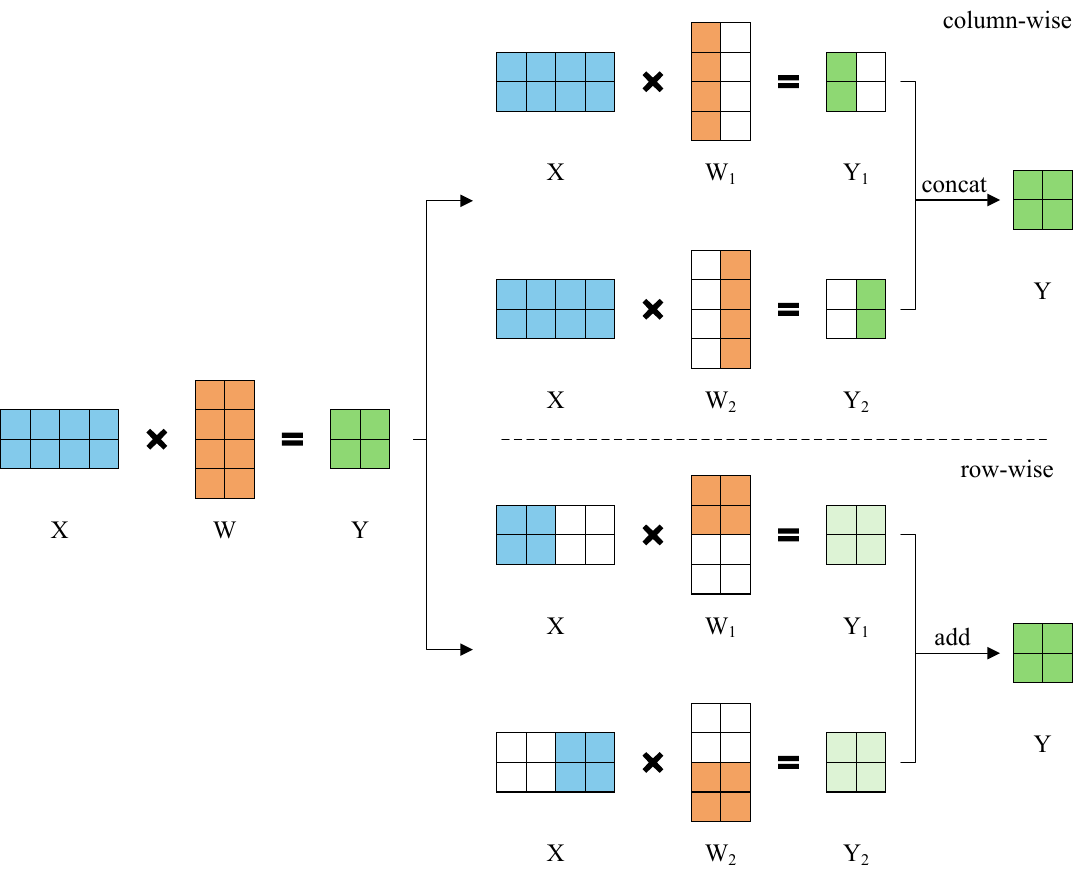}
\caption{Splitting of matrix multiplication.}
\label{fig:gemm}
\end{figure}

Matrix multiplication can be divided into smaller operations by row or column, allowing the results to be calculated separately and then aggregated to obtain the final outcome, as shown in Fig. \ref{fig:gemm}. In line with this idea, Megatron-LM \cite{shoeybi_megatronlm_2020} has developed a tensor parallelism approach specifically designed for the transformer architecture. Taking the MLP layer as an example, it consists of two parameter matrices: $W_1$ with a shape of $(h, 4h)$ and $W_2$ with a shape of $(4h, h)$, here we denote hidden size as $h$. Initially, $W_1$ is partitioned column-wise, while $W_2$ is partitioned row-wise. This results in the parameter matrices stored on each device having the shapes: $W_1^\prime(h, 4h/N_d)$ and $W_2^\prime(4h/N_d, h)$. After forward pass on each device is completed, an all-reduce is performed to obtain the final result.

Megatron-LM TP divides tensors into one-dimensional segments. While this approach reduces model parameters stored on each device to $\frac {1}{N_d}$, it still needs to store complete intermediate activations on each device. Additionally, forward and backward computations of each transformer layer involve 4 all-reduce, causing significant communication overhead. To address this, researchers have proposed multidimensional tensor partitioning methods \cite{xu_efficient_2023a, wang_tesseract_2022, bian_maximizing_2021}, which further reduce memory overhead on each device and mitigate the frequent communication issues of 1D TP \cite{li_colossalai_2023}. In practice, TP is typically applied within nodes rather than across nodes to prevent communication from becoming a bottleneck for model training efficiency \cite{korthikanti_reducing_2023}.

\item \textbf{Pipeline Parallelism}

Pipeline parallelism, or inter-layer parallelism, involves partitioning the model layers horizontally into multiple stages. Each stage is assigned to different devices for sequential execution. 
PP not only reduces memory demands and computational workloads on each device but also has less communication overhead. However, a major challenge associated with PP is device idle time (often referred to as pipeline bubbles). While asynchronous PP schemes \cite{harlap_pipedream_2018, yang_pipemare_2021} can eliminate bubbles, they have no guarantee of model convergence \cite{qi_pipeline_2024}. As synchronous PP schemes continue to improve, their efficiency are comparable with asynchronous PP, making them a more favorable option. 

\begin{figure*}[!t]  
    \centering
    \subfloat[Pipeline schedules (top: 1F1B, mid: interleaved 1F1B, bottom: ZB-H1, $D_i$ represents device $i$)]{\includegraphics[width=.67\textwidth]{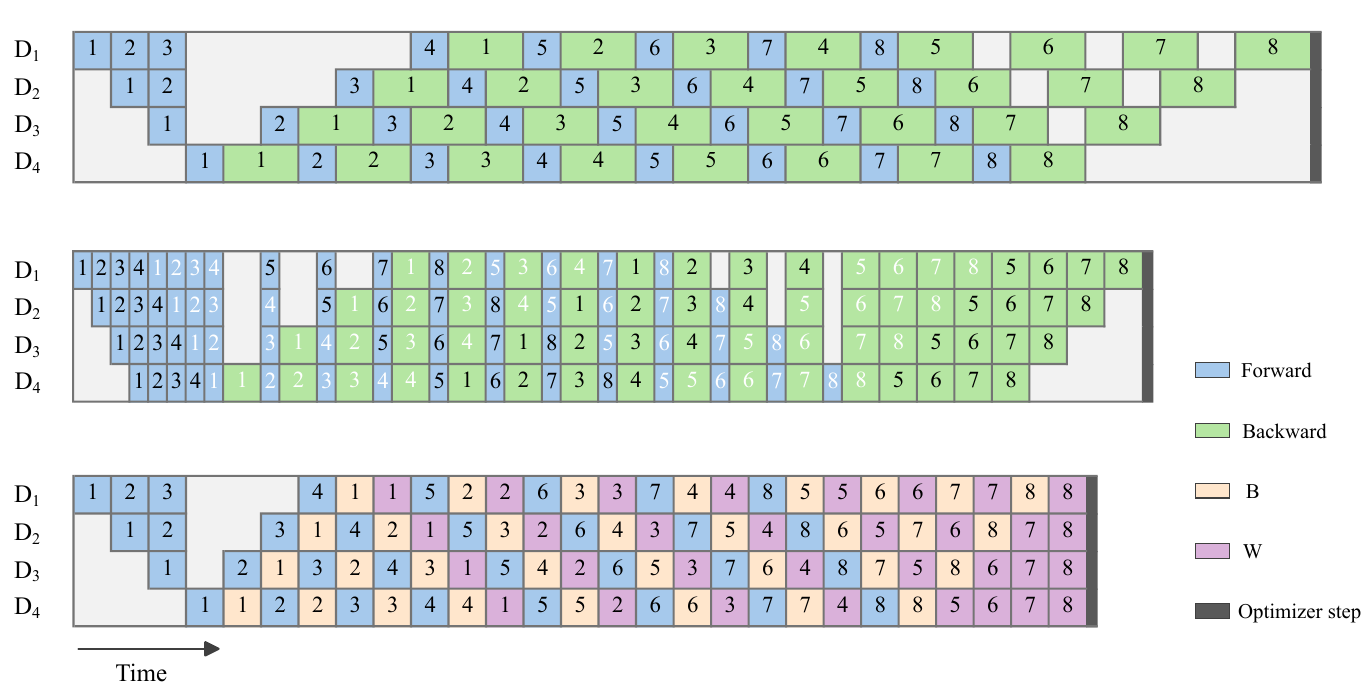}\label{fig:1f1b_pipeline_schedule}}
    \hspace{5pt}
    \subfloat[Device placements (left: 1F1B / ZB-H1, right: interleaved 1F1B)]{\includegraphics[width=.3\textwidth]{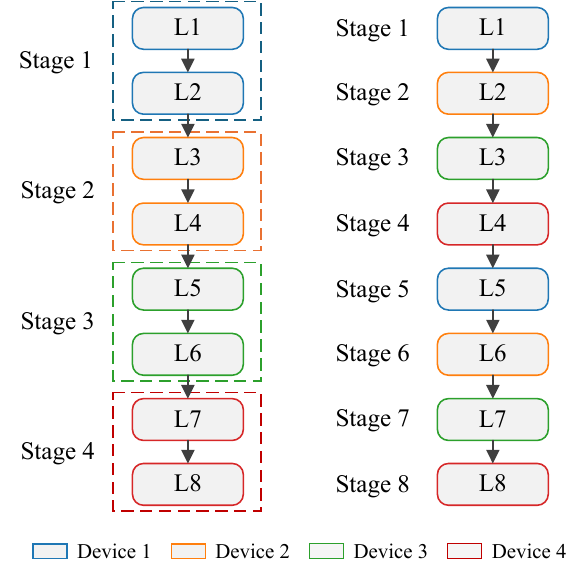}\label{fig:1f1b_device_placement}}
    \caption{Pipeline schedules and device placements of 1F1B, interleaved 1F1B, and ZB-H1. Interleaved 1F1B partitions model into finer-grained stages and assign each device with multiple stages to reduce pipeline bubbles.}
    \label{fig:1f1b}
\end{figure*}

In vanilla PP, data can be passed to the next device only after the preceding device completes its calculation. This leads to a waste of resources and diminishes system efficiency. GPipe \cite{huang_gpipe_2019} splits each input batch into several micro-batches. After completing the processing of each micro-batch, each device immediately passes its results to the next device. GPipe significantly reduces pipeline bubbles, 
however, it faces the issue of high activation memory consumption.

Let $N_s$ denote the number of stages in the model, $N_d$ denote the number of devices, and $N_m$ denote the number of micro-batches. In GPipe, each device must retain activations of $N_m$ micro-batches. To reduce memory consumption, researchers introduced the 1F1B schedule \cite{harlap_pipedream_2018, narayanan_efficient_2021a, fan_dapple_2021a} (see Fig. \ref{fig:1f1b_pipeline_schedule}). This schedule effectively reduces the number of on-the-fly micro-batches by executing the backward pass in advance. Its pipeline bubble is equal to that of GPipe, and the activations of micro-batches required to store on each device decrease from $N_m$ to $N_d$. 

The aforementioned PP schemes assume that $N_s=N_d$, with each device possessing one stage. Narayanan et al. \cite{narayanan_efficient_2021a} split the stages, allowing each device to have multiple stages (we demonstrate an example in Fig. \ref{fig:1f1b_device_placement}). This interleaved 1F1B schedule (Fig. \ref{fig:1f1b_pipeline_schedule}) performs finer-grained scheduling, reducing pipeline bubbles and enhancing efficiency. However, the trade-off for this includes increased communication and higher activation memory usage.

The computational cost of backpropagation is greater than that of forward pass, resulting in longer computation time, causing the existence of pipeline bubbles. Qi et al. \cite{qi_zero_2024} suggest splitting the backward pass into calculating the gradients of activations and the gradients of parameters (denote the two calculations as $B$ and $W$ respectively). Theoretically, one could complete $B$ passes of all layers before performing $W$ passes for each layer. Based on this finding, they proposed ZB-H1 schedule, which offers the flexibility to schedule $W$ after the corresponding $B$ within the same stage, effectively filling bubbles in the schedule (depicted in Fig. \ref{fig:1f1b_pipeline_schedule}). 

PP typically has the issue of high activation memory usage, and the 1F1B schedules mentioned above have the problem of memory imbalance. Qi et al. \cite{qi_pipeline_2024} proposed an analytic framework to decompose existing pipeline schedules into building blocks and found that peak memory usage of a pipeline schedule is highly related to the lifespan of the building blocks. 

Compared to TP, PP only communicates between adjacent stages and has lower communication overhead. Therefore, it is usually applied between nodes to scale up model size \cite{narayanan_efficient_2021a}.

\item \textbf{Sequence Parallelism}

Supporting long sequence lengths is crucial for applying AI to scientific endeavors. Many scientific challenges inherently involve complex analysis of large-scale, high-dimensional data, frequently presented as long sequences. Long-sequence issues are prevalent in fields including structural biology, chemistry, drug development, and atmospheric science. For instance, many proteins consist of hundreds or even thousands of amino acid residues, where the sequence context of each residue can be essential to its 3D structure. Long-sequence support enhances the ability to capture non-local interactions among residues, thereby improving the quality of predictions. Additionally, complex molecules or chemical reactions often necessitate long sequence representations, including molecular formulas, SMILES, reaction pathways, etc. Climate prediction models require analyzing long-term historical and global spatial data, represented in time series or grid formats. Despite the importance of long-sequence support, processing long sequences demands significant increases in computational power and memory, with the memory usage of attention mechanisms rising quadratically with sequence length.

In traditional distributed training techniques such as DP and TP, model parameters or input data are partitioned and assigned to multiple devices for processing. However, when the model needs to handle long inputs, these methods may lose efficiency due to memory bottlenecks or computational overhead. Sequential parallel (SP) is relatively novel and differs from DP and MP methods in that it aims to reduce the memory overhead caused by activations during model training. Although SP strategies divide input along sequence dimension, the specific implementation varies.

\begin{figure*}[!t]  
    \centering
    \includegraphics[width=1\textwidth]{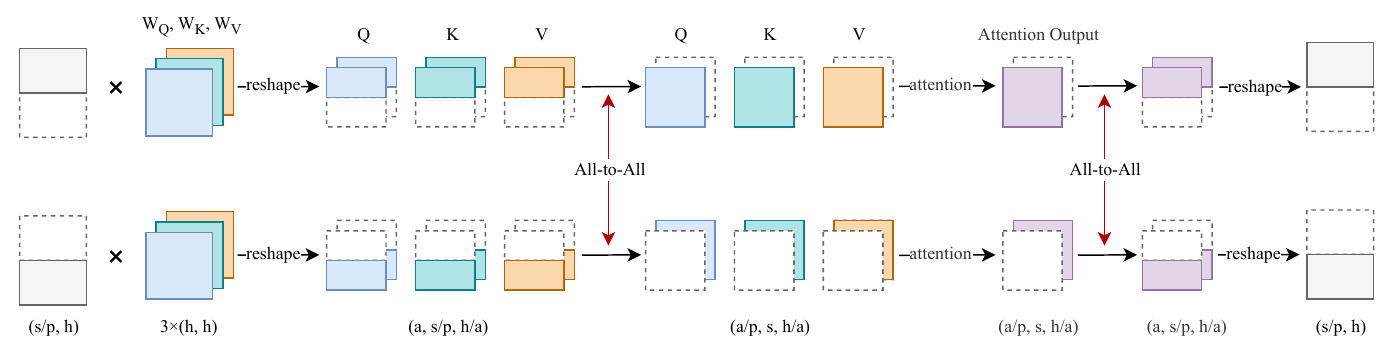}
    \caption{Workflow of Ulysses sequence parallelism with 2 devices. Notations: $s$ for sequence length, $h$ for hidden size, $a$ for number of attention heads, and $p$ for SP degree.}
    \label{fig:ulysses}
\end{figure*}

\begin{figure}[t]  
    \centering
    \includegraphics[width=1\linewidth]{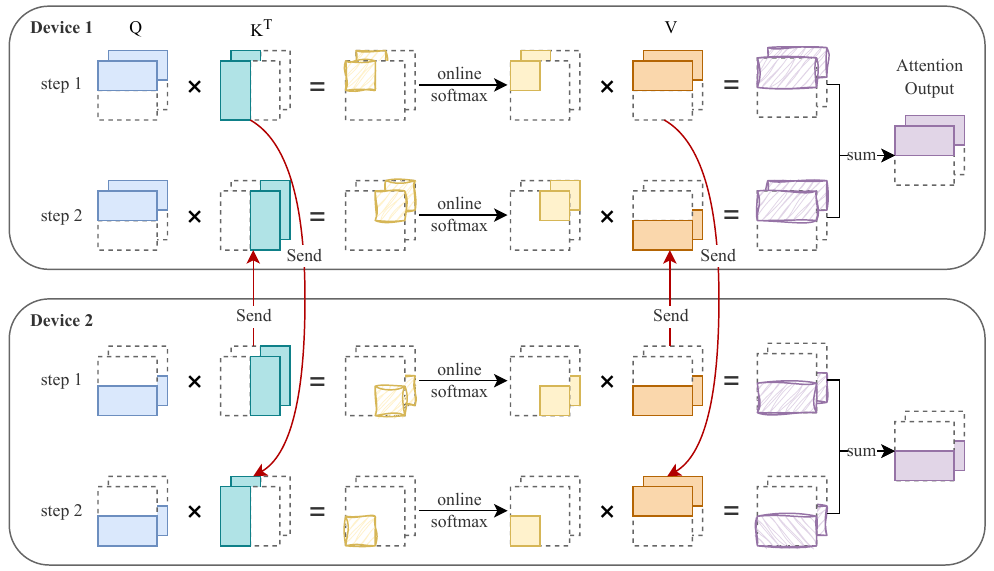}
    \caption{Workflow of Ring Attention sequence parallelism with 2 devices.}
    \label{fig:ring-attention}
\end{figure}

Li et al. \cite{li_sequence_2023} considered solving the problem from a system perspective by combining ring-style communication with self-attention, allowing the input sequence to be distributed to each device and calculating attention scores across devices. Compared with Megatron-LM TP \cite{shoeybi_megatronlm_2020}, the proposed method is not limited by the number of attention heads, as long as the sequence length is divisible by the degree of SP.

In Megatron-LM TP, some activations within the transformer layer are not distributed across devices, leading to increased memory overhead. The SP method proposed by Li et al. \cite{li_sequence_2023} can help mitigate this issue; however, it requires replicating model parameters and optimizer states across all devices, making it impractical for large-scale model training. Korthikanti et al. \cite{korthikanti_reducing_2023} found that the LaterNorm and Dropout operations within the transformer layer operate independently along the sequence dimension. Consequently, modifications were implemented based on Megatron-LM to effectively distribute the computational workloads and activations associated with LaterNorm and Dropout, without incurring additional communication overhead.

Jacobs et al. proposed Ulysses \cite{jacobs_system_2024}, a method that divides the input along sequence dimension onto all devices and conducts all-to-all communication on the Query (Q), Key (K), and Value (V) matrices before performing attention computation. This ensures that each device operates with disjoint attention heads, allowing for distributed attention calculation. Once the calculations are complete, another all-to-all is performed to get the results and restore the state that inputs were divided along sequence dimension. This process is presented in Fig. \ref{fig:ulysses}. Ulysses SP and Megatron-LM TP share similarities in the distributed calculation of self-attention, both concerning limitations on the number of attention heads. A key advantage of Ulysses is that the communication volume for a single device decreases linearly as SP degree increases, while in Megatron-LM, it is independent of TP degree.

Liu et al. \cite{liu_ringattention_2023} developed a ring-style distributed attention computation method that optimizes resource usage by conserving a portion of the Q, K, and V blocks. During calculations, each device exchanges K and V blocks with its neighboring devices, enabling it to perform computations using blockwise self-attention and feedforward networks (FFN) \cite{liu_blockwise_2023}, as depicted in Fig. \ref{fig:ring-attention}. In contrast to the approach taken by Li et al. \cite{li_sequence_2023}, this method can eliminate communication overhead by picking a proper chunk size and overlapping communication with computation.
\end{enumerate}

\subsubsection{Offloading}\label{sec:Offloading}
Offloading refers to the process of transferring specific storage tasks from GPU/TPU memory to main memory (CPU memory) or lower-cost storage devices, such as NVMe SSD. This technique alleviates memory overhead through increased data communication and enhances the utilization of hardware resources.

SwapAdvisor \cite{huang_swapadvisor_2020} takes the data flow graph of a model as input and selects legitimate memory allocation and operator scheduling based on the graph. By precisely planning which and when tensors require swapping in and out, SwapAdvisor achieves optimal overlap between computation and communication. The integration of genetic algorithms enables the exploration of the vast search space for finding the optimal combination of memory allocation and operator scheduling. In contrast to previous heuristic-based swapping methods \cite{rhu_vdnn_2016a, le_tflms_2019}, SwapAdvisor significantly enhances memory efficiency and computational performance through its dual optimization approach. Notably, due to the lack of consideration of GPU communication, SwapAdvisor has limitations in multi-GPU training scenarios.

Ren et al. \cite{ren_zerooffload_2021} introduced an offloading strategy designed for mixed precision training using the Adam optimizer, which allows for seamless integration with MP. They offload all FP32 model states and FP16 gradients to CPU memory and utilize CPU to compute parameter updates while FP16 model parameters remain on GPU to perform forward and backward passes. It enables the overlapping of swapping processes with computation, mitigating most communication overheads. Furthermore, this approach can be combined with ZeRO-2 DP to achieve good scalability.

ZeRO-infinity \cite{rajbhandari_zeroinfinity_2021} enhances the capabilities of ZeRO-3 by adding support for heterogeneous memory, thereby leveraging the entire spectrum of memory types within the system (GPU, CPU, and SSD) to enable large-scale model training. By employing the Infinity Offload Engine to offload model states to NVMe storage and efficient communication optimizations, ZeRO-infinity enables super-linear scalability of training throughput when scaling from 64 GPUs to 512 GPUs for training a 1T-parameter model, the overall throughput increased from 2.8 petaflops to 25 petaflops, surpassing linear scaling. ZeRO-Infinity trains 500B model with throughput comparable to 3D-parallelism, if increasing the model size to 1T, 3D-parallelism runs out of memory, while ZeRO-Infinity still reaches over 70\% of achievable hardware FLOPS utilization. On 32 Nvidia V100 DGX-2 nodes, ZeRO-infinity supports Transformer models with up to 32T parameters, which is 50× the maximum size accommodated by 3D-parallelism (around 650B parameters).

A comparative summary of representative offloading strategies, including SwapAdvisor, ZeRO-Offload, and ZeRO-Infinity, is provided in Table \ref{tab:overview-offloading}, highlighting their core features, applicable scenarios, and performance.

\begin{table*}[!t]
\centering
\caption{An overview of offloading methods}
\label{tab:overview-offloading}
\resizebox{\linewidth}{!}{
    \begin{tabular}{clll}
    \Xhline{1px}
    \textbf{Method} & \multicolumn{1}{c}{\textbf{Feature}} & \multicolumn{1}{c}{\textbf{Applicable Scenarios}} & \multicolumn{1}{c}{\textbf{Performance}} \\ \hline
    SwapAdvisor & \begin{tabular}[c]{@{}l@{}}- Optimize scheduling, memory allocation\\ ~~and swap planning simultaneously\end{tabular} & - Single-GPU training & \begin{tabular}[c]{@{}l@{}}Achieve 53\% - 99\% of the training throughput of the\\  ideal baseline\end{tabular} \\ \hline
    ZeRO-Offload & - Utilize CPU to perform weight update & \begin{tabular}[c]{@{}l@{}}- Apply in conjunction with ZeRO-2 and\\ ~~mixed precision\end{tabular} & \begin{tabular}[c]{@{}l@{}}Outperforms SwapAdvisor by 23\% in training throughput;\\ Enables training 1.62× larger model (vs. SwapAdvisor) on a\\ single GPU, and increases model scale by 4.5× (vs. Megatron)\\  and 7.8× (vs. ZeRO-2) on a single DGX-2 node\end{tabular} \\ \hline
    ZeRO-Infinity & - Exploit CPU and NVMe memory & \begin{tabular}[c]{@{}l@{}}- Apply in conjunction with ZeRO-3\\ - Train extremely large models\\ ~~(e.g., \textgreater{}1T params)\end{tabular} & Enable 1T model training on a single DGX-2 node without MP \\
    \Xhline{1px}
    \end{tabular}
}
\end{table*}

\subsection{Hardware-Software Co-Optimization}\label{sec:Hardware}
As transformer models scale, training them efficiently requires both algorithmic innovations and tight coupling between hardware characteristics and software design. By aligning software execution patterns with hardware capabilities—such as memory hierarchy, bandwidth, and parallelism—these techniques significantly improve training throughput and memory efficiency.

A representative example of hardware-software co-optimization in training transformers is FlashAttention \cite{dao_flashattention_2022}, which re-architects the attention computation to exploit GPU memory hierarchy, achieving significant speed and memory efficiency improvements. It introduces an IO-aware exact attention algorithm that significantly improves memory efficiency by minimizing expensive reads/writes between high-bandwidth GPU memory (HBM) and on-chip SRAM. Through tiling and recomputation, FlashAttention avoids materializing large intermediate attention matrices, achieving up to 3.5× training time speedup and reducing memory consumption of self-attention from $O(n^2)$ to $O(n)$, which enables scaling transformer models to longer sequences.

FlashAttention-2 \cite{dao_flashattention2_2023} builds upon FlashAttention by optimizing GPU work partitioning and parallelism. It reduces non-matrix-multiply operations, increases GPU occupancy by parallelizing across sequence dimensions, and minimizes shared memory communication. As a result, FlashAttention-2 achieves 2× speedup over FlashAttention and reaches up to 73\% of the theoretical peak FLOPS on A100 GPU. Focusing on hardware-aware optimizations for Hopper architecture, FlashAttention-3 \cite{shah_flashattention3_2024} optimizes attention computation on GPUs using asynchronous execution, interleaved GEMM-softmax operations, and FP8 low-precision with block quantization. It achieves 1.5–2.0× speedup over FlashAttention-2 (85\% utilization in BF16, 1.3 PFLOP/s in FP8) while reducing FP8 error by 2.6× compared to baseline FP8 attention.

MPress \cite{zhou_mpress_2023} is a novel system designed to break the GPU memory wall for billion-scale model training on multi-GPU servers. It combines inter-operator parallelism with a new D2D swap technique to efficiently transfer tensors between GPUs, leveraging high-bandwidth NVLink connections. MPress dynamically selects the best memory-saving strategies based on tensor live intervals and available spare memory. Evaluations show that MPress can train larger models compared to recomputation baseline (3.7× for BERT and 1.7× for GPT) while maintaining high training throughput, and achieves 1.4–2.3× speedups compared to ZeRO-Series baselines \cite{ren_zerooffload_2021, rajbhandari_zeroinfinity_2021} under the same memory reduction.

Beyond these optimizations, broader efforts in the LLM community have also explored general-purpose hardware–software co-design frameworks. Wattanawong et al. \cite{Wattanawong:EECS-2023-92} propose an approach that jointly evolves Transformer architectures and hardware accelerators via neural architecture search. Their framework achieves up to 10× improvement in energy-delay product (EDP) with minimal perplexity degradation, demonstrating that hardware-aware architectural adaptations can substantially reduce inference costs. Meanwhile, Guo et al. \cite{guo_survey_2025} present a comprehensive survey of LLM-centric co-design, outlining the challenges posed by large model size, energy consumption, and memory bottlenecks.

\section{Tailored Memory-Efficient Training Techniques: A Case Study}\label{sec:Tailored Memory-Efficient Training Techniques: A Case Study}
OpenFold \cite{ahdritz_openfold_2024} is an open-source implementation of AlphaFold 2, which has a prediction quality that matches AlphaFold 2 and runs faster on most proteins with less memory usage. This improves the ease and performance of training new models and performing large-scale predictions. OpenFold has made various optimizations to AlphaFold 2's training schedule. OpenFold was trained on a cluster of 44 NVIDIA A100 GPUs, each with 40GB of memory. It enables FP16 mixed precision training mode to enhance training speed while minimizing memory usage. To reproduce the effective batch size utilized during AlphaFold 2 training as closely as possible, OpenFold implements 3-way gradient accumulation. In terms of distributed training, Deepspeed \cite{rasley_deepspeed_2020a} and ZeRO stage 2 \cite{rajbhandari_zero_2020b} are employed to distribute the optimizer states across each GPU, thus reducing model redundancy. Additionally, they refactored the model by replacing element-wise operations with in-place equivalents wherever possible to minimize unnecessary memory allocation. Advanced implementation of the self-attention \cite{rabe_selfattention_2022, dao_flashattention_2022} was also adopted to lower memory usage and speed up attention computation. Furthermore, they optimize CUDA kernels for certain model modules, decreasing memory overhead.

Despite its impressive success in prediction accuracy, AlphaFold 2's computational and memory costs during training are much higher than those of vanilla transformers, and its architecture is computationally inefficient on GPUs. To address these issues, Cheng et al. proposed FastFold \cite{cheng_fastfold_2024}, an efficient implementation of AlphaFold 2. FastFold introduces diverse LLM training techniques, significantly reducing the cost of training and inference for AlphaFold 2 models. 

Unlike many existing LLMs, AlphaFold 2's architecture derives its main memory overhead from intermediate activations, requiring more than 20 GB memory in each attention module \cite{cheng_fastfold_2024}, which cannot be well distributed using common methods like ZeRO DP, PP, or TP. Therefore, FastFold introduces a model parallel approach that focuses on reducing activations to optimize AlphaFold 2 training. Dynamic Axial Parallelism (DAP) is a technique tailored for the EvoFormer modules in AlphaFold 2. It retains complete model parameters on each device while partitioning inputs and intermediate activations across devices. The MSA representation and Pair representation within the EvoFormer module contain two distinct sequence dimensions, with computations performed along only one of these dimensions at a time. Consequently, DAP considers partitioning the other sequence dimension not in use. Due to the alternating calculations across different sequence dimensions, all-to-all communication is necessary to ensure that each device possesses the required data for computation. This design allows for the distribution of activations among devices, significantly reducing the training memory overhead of AlphaFold 2 and enhancing both efficiency and scalability. When the model is trained with DAP on a considerable number of devices,  throughput can be improved by disabling gradient checkpointing, as DAP has saved enough memory. Compared with the original AlphaFold 2, FastFold has reduced total training time from 11 days to 2.81 days, representing a reduction of 3.91×, and compared with OpenFold, it has improved by 2.98×.

Although OpenFold utilizes advanced systematic approaches for performance and memory optimization, the pre-training cost of AlphaFold 2 remains high. While AlphaFold 2 has a relatively small scale with only 93M parameters, its activation memory usage during training is extremely high. Song et al. \cite{song_deepspeed4science_2023} designed customized attention kernels for the four types of attention variants in the Evoformer module to save memory. This approach lowers peak memory consumption by a factor of 13 compared to the OpenFold implementation. When combined with DAP, it can further reduce memory overhead.

Zhu et al. \cite{zhu_scalefold_2024a} conducted a comprehensive analysis of AlphaFold 2's training performance and identified that its poor scalability is primarily due to communication imbalance and inefficient computation. To address these issues, they combined DAP with various optimization methods, such as designing non-blocking data pipelines and performing operator fusion, to construct ScaleFold. ScaleFold allows for the training of AlphaFold 2 on 2080 NVIDIA H100 GPUs and shortens the pre-training time to merely 10 hours.

\section{Future Trends and Challenges}\label{sec:Future Trends and Challenges}
\begin{itemize}[wide]
    \item \textbf{Scaling transformer models remains both a challenge and an opportunity.} The number of parameters and the data processing capabilities of transformer models play a crucial role in their ability to address complex scientific problems. However, the memory demands for training these models often surpass current hardware limitations, creating a bottleneck in scaling. Optimizing memory usage during training not only enhances efficiency but also provides possibilities for exploring even larger models. This optimization is vital for scientific models, where tasks like protein structure prediction, molecular generation, and climate prediction demand models can handle high-dimensional and long-sequence data while scaling up models can enhance the accuracy of these applications. Moreover, employing effective memory optimization strategies allows practitioners to train models using fewer GPUs, resulting in substantial savings in cost and energy. It lowers the barriers to training large transformer models, promoting greater participation from researchers in scientific fields.

    \item \textbf{Advanced memory optimization toolkits are available, but underutilized in science-focused LLM work.} For the training of transformer-based scientific large models, there exists diverse memory optimization methods and training libraries, such as DeepSpeed, Megatron-LM, Colossal-AI, FairScale, Hugging Face Accelerate, etc., most of which support various optimization techniques introduced in this survey. However, our investigation indicates that many studies employing LLMs for scientific research have not fully leveraged these memory-efficient training approaches. Thus, there remains considerable room for optimization in LLM training in scientific tasks. This survey aims to systematically summarize the memory optimization techniques, hoping to promote their application in scientific domains and assist researchers in fully harnessing the capabilities of LLMs to accelerate scientific discovery and innovation.

    \item \textbf{The optimization of transformer variants in science require customized optimization strategies.} While many memory optimization techniques are effective for standard transformers, they may not generalize to modules like: Evoformer in AlphaFold 2, SE(3)-Transformer in RoseTTAFold, and Pairformer in AlphaFold 3, which deviate significantly in computational complexity and memory behavior. Optimizing these models often requires domain expertise to develop tailored strategies that align with the specific structure of the model. This process necessitates not only a profound understanding of the characteristics of corresponding scientific tasks but also the co-optimization of software and hardware to achieve optimal performance. Designing effective memory optimization methods for these models remains an urgent challenge that needs to be solved.

    \item \textbf{Memory-efficient training is an essential direction for future research.} As model size and data scale continue to grow, memory-efficient training methods have become a key factor for advancing transformer models for science. It is crucial to integrate memory optimization techniques with advanced training frameworks, develop customized optimization strategies, and further lower the resource requirements for model training. Furthermore, future research could focus on creating more versatile guidelines for designing optimization strategies and promoting the widespread use of high-performance computing resources in scientific fields through collaborative software and hardware design.
\end{itemize}

\section{Conclusion}\label{sec:Conclusion}
In scientific fields, the performance and complexity of large transformer models have continually evolved as their applications expand into fields including biology, medicine, chemistry, and meteorology. However, their training often incurs significant memory demands and computational overhead, posing tough challenges to existing hardware resources. Memory-efficient training methods not only enhance training efficiency but also provide support for exploring the potential of larger-scale models in AI for science. This survey provides a systematic summary of the memory optimization techniques that have gained popularity in recent years for training transformers. Additionally, we use AlphaFold 2 as an illustrative example to introduce customized memory-efficient training approaches tailored to specific task requirements, serving as a beneficial reference for researchers interested in applications of transformer-based models for scientific tasks.

While existing memory optimization techniques and training frameworks for transformers have made significant strides in NLP, their application in scientific fields has yet to achieve widespread adoption. Scientific models often encounter specific task requirements, including long-sequence input and multimodal data, which can result in high activation memory overhead, making it crucial to choose appropriate methods for memory optimization. 




\bibliographystyle{fcs}
\bibliography{ref}

\end{document}